\documentclass[twoside]{article}
\usepackage[accepted]{aistats2e}

\usepackage{mdwlist,color}
\usepackage{graphicx} 
\usepackage{multirow}
\usepackage{algorithm}
\usepackage{algorithmic}
\usepackage[tight]{subfigure}
\usepackage{wrapfig}
\usepackage{times}
\usepackage{natbib}

\usepackage[compact]{titlesec}
\titlespacing{\section}{0pt}{2ex}{1ex}
\titlespacing{\subsection}{0pt}{2ex}{1ex}

\usepackage{Definitions}

\usepackage{mathrsfs}
\usepackage{color}
\newcommand{\Le}[1]{{\color{blue}{\bf\sf [Le: #1]}}}
\newcommand{\Lee}[1]{{\color{red}{\bf\sf [DB: #1]}}}

\newcommand{\hCcal}{\widehat{\Ccal}}
\newcommand{\tCcal}{\widetilde{\Ccal}}
\newcommand{\hUcal}{\widehat{\Ucal}}
\newcommand{\tUcal}{\widetilde{\Ucal}}

\newcommand{\commentout}[1]{}

\begin{document} 

\twocolumn[
\aistatstitle{Kernel Belief Propagation}
\vskip -0.25in
\aistatsauthor{Le Song,$^1$ Arthur Gretton,$^{1,2}$ Danny Bickson,$^1$ Yucheng Low,$^1$ Carlos Guestrin$^1$}
%
\aistatsaddress{$^1$ School of Computer Science, CMU; $^2$Gatsby Computational Neuroscience Unit \& MPI for Biological Cybernetics}

\vskip -0.2in
]


\begin{abstract}    
\vspace{-3mm}
We propose a nonparametric generalization of belief propagation, Kernel Belief Propagation (KBP), for pairwise Markov random fields. Messages are represented as functions in a reproducing kernel Hilbert space (RKHS), and message updates are simple linear operations in the RKHS.  KBP makes none of the assumptions commonly required in classical BP algorithms: the variables need not arise from a finite domain or a Gaussian distribution, nor must their relations take any particular parametric form. Rather, the relations between variables are represented implicitly, and are learned nonparametrically from training data. KBP has the advantage that it may be used on any domain where kernels are defined ($\mathbb{R}^d$, strings, groups), even where explicit parametric models are not known, or closed form expressions for the BP updates do not exist. 
The computational cost of message updates in KBP is polynomial in the training data size. We also propose a constant time approximate message update procedure by representing messages using a small number of basis functions. In experiments, we apply KBP to image denoising, depth prediction from still images, and protein configuration prediction: KBP is faster than competing classical and nonparametric approaches (by orders of magnitude, in some cases), while providing significantly more accurate results.

\end{abstract}


\vspace{-6mm}
\section{Introduction}
\vspace{-3mm}

Belief propagation is an inference algorithm for graphical models that has been widely and successfully applied in a great variety of domains, including vision~\citep{SudIhlFreWil03},
protein folding~\citep{YanWei02}, and turbo decoding~\citep{MceMacChe98}.
In these applications, the variables are usually assumed either to be
finite dimensional, or in continuous cases, to have a Gaussian distribution \citep{WeiFre01}. 
In many applications of graphical models, however, the variables of interest are 
naturally specified by continuous, non-Gaussian distributions.
For example, in constructing depth maps from 2D images, the depth is both continuous valued and has a multimodal distribution. Likewise, in protein folding, angles are modeled as continuous valued random variables, and are predicted from amino acid sequences. 
In general, 
multimodalities, skewness, and other non-Gaussian statistical features are present in a great many real-world problems. 
The corresponding inference procedures for
parametric models typically involve integrals for which
no closed form solutions exist,
and are without  computationally tractable exact message updates. 
Worse still,   parametric models for the relations between the 
variables may not even be known, or may be prohibitively complex.

Our first contribution in this paper is a novel  generalization of belief propagation for pairwise Markov random fields, {\em Kernel BP}, based on a reproducing kernel Hilbert space (RKHS) representation of the relations between  random variables. This extends earlier work of \citet{SonGreGue10} on inference for trees to the case of graphs with loops. The algorithm consists of two parts, both nonparametric: 
first, we {\em learn} RKHS representations of the relations between variables directly
 from training data, which removes
the need for an explicit parametric model. Second, we propose 
a belief propagation algorithm for {\em inference} based on these learned relations,  
where each update is a linear operation in the RKHS (although the relations themselves may be highly nonlinear in the original space of the variables). 
Our approach applies not only to  continuous-valued non-Gaussian variables, but also generalizes to strings and graphs \citep{SchTsuVer04}, groups \citep{FukSriGreSch09}, compact manifolds \citep[][Chapter 17]{Wendland05}, and other domains on which kernels may be defined.


A number of alternative approaches  have been developed to perform inference in the continuous-valued non-Gaussian setting. 
 \cite{SudIhlFreWil03}~proposed an approximate belief propagation algorithm for pairwise Markov random fields, where the parametric forms of the node and edge potentials are supplied in advance, and the messages are approximated as mixtures of Gaussians: we refer to this approach as {\em Gaussian Mixture BP} (this method was introduced as ``nonparametric BP'', but it is in fact a Gaussian mixture approach).
Instead of mixtures of Gaussians, \cite{IhlMcA09} used particles to approximate the messages, resulting in the {\em Particle BP} algorithm. 
Both Gaussian mixture BP and particle BP assume the potentials to be pre-specified by the user: the methods described are purely approximate message update procedures, and do {\em not} learn the model from training data. 
By contrast, kernel BP 
learns the model, is computationally tractable even before approximations are made, and leads to an entirely different message update formula than the Gaussian Mixture and Particle representations. 

A direct implementation of kernel BP has a reasonable computational cost:
each message update  costs $O(m^{2}d_{\mathrm{max}})$ when computed exactly, where $m$ is the number of training examples and $d_{\mathrm{max}}$ is the maximum
degree of a node in the graphical model. 
For massive data sets and numbers of nodes, as occur in
image processing, this cost might still be expensive.
Our second contribution is
a novel {\em constant time} approximate message update procedure, where we express
the messages in terms of a small number $\ell\ll m$ of representative RKHS basis functions learned from training data. Following an initialization cost linear in $m$,  the cost per message update is decreased to $O(\ell^{2}d_{\mathrm{max}})$, independent of the number  of training points $m$. 
Even without these approximate constant time updates, kernel BP is substantially faster than Gaussian mixture BP and particle BP. Indeed, an exact implementation of Gaussian mixture BP would have an exponentially increasing computational and storage cost with number of iterations.  In practice, both Gaussian mixture and particle BP require a Monte Carlo resampling procedure at every node of the graphical model.
%

Our third contribution is a thorough evaluation of kernel BP against other nonparametric BP approaches. We apply both kernel BP and competing approaches to an image denoising problem, depth prediction from still images, protein configuration prediction, and paper topic inference from citation networks: these are all large-scale problems, with continuous-valued or structured random variables  having complex underlying probability distributions.  In all cases, kernel BP performs outstandingly, being orders of magnitude faster than both Gaussian mixture BP and particle BP, and returning more accurate results.

\vspace{-3mm}
\section{Markov Random Fields And Belief Propagation}
\vspace{-3mm}

\setlength{\abovedisplayskip}{1pt}
\setlength{\abovedisplayshortskip}{0pt}
\setlength{\belowdisplayskip}{1pt}
\setlength{\belowdisplayshortskip}{0pt}
\setlength{\jot}{0pt}

We begin  with a short introduction to pairwise Markov random fields (MRFs) and the belief propagation algorithm. 
A pairwise Markov random field (MRF) is defined on an undirected graph 
$\Gcal:=(\Vcal,\Ecal)$
with nodes $\Vcal:=\{1,\ldots,n\}$ connected by edges
in $\mathcal{E}$. Each node
$s\in\Vcal$ is associated with a random variable $X_{s}$ on the 
domain
$\mathcal{X}$ (we assume a common domain for ease of notation, but in practice the domains can be different), and $\Gamma_{s}:=\cbr{t|(s,t)\in\Ecal}$ is the set of neighbors of node $s$ with size $d_{s}:=\left|\Gamma_{s}\right|$. In a pairwise MRF, the joint distribution of the variables $\bm{X}:=\{X_1,\ldots,X_{|\Vcal|}\}$ is assumed to factorize according to a model $\PP(\bm{X})=\frac{1}{Z} \prod_{(s,t)\in\mathcal{E}}\Psi_{st}(X_{s},X_{t})\prod_{s\in\mathcal{V}}\Psi_{s}(X_{s})$, where $\Psi_{s}(X_{s})$ and $\Psi_{st}(X_{s},X_{t})$ are node and edge potentials respectively, and $Z$ is the partition function that normalizes the distribution.  

The inference problem in an MRF
is defined as calculating the marginals $\PP(X_s)$ for nodes $s\in\Vcal$ and $\PP(X_s,X_t)$ for edges $(s,t)\in\Ecal$. The marginal
$\PP(X_s)$ not only provides a measure of uncertainty of $X_s$, but also 
leads to a point estimate $x^\star_s:=\argmax \PP(X_s)$. 
Belief Propagation (BP) is an iterative algorithm for performing inference in MRFs~\citep{Pearl88}. BP represents intermediate results of marginalization steps as messages passed between adjacent nodes: a message $m_{ts}$ from $t$ to $s$ 
is calculated based on  
messages $m_{ut}$ from all neighboring nodes $u$ of $t$ besides $s$,~i.e., 
\begin{align}    
  \hspace{-3mm}
  m_{ts}(X_{s})=\int_{\Xcal}\Psi_{st}&(X_{s}, X_{t})\Psi_{t}(X_{t})
  \prod_{u\setminus s}m_{ut}(X_{t})dX_t.
  \label{eq:classicalBP}  
\end{align}
Note that we use $\prod_{u\setminus s}$ to denote $\prod\nolimits_{u\in\Gamma_{t}\setminus s}$, where it is understood that the indices range over all neighbors $u$ of $t$ except $s$. 
This notation also applies to operations other than the product. 
The update in~\eq{eq:classicalBP}~is iterated across all nodes until a fixed point, $m_{ts}^\star$, for all messages is reached. The resulting
node beliefs (estimates of node marginals) are given by
$
  \BB(X_{s})\propto\Psi_{s}(X_{s})\prod\nolimits_{t\in\Gamma_{s}}m_{ts}^\star(X_{s}).
$

For acyclic or tree-structured graphs, BP results in
 node beliefs $\BB(X_s)$ that converge to the node marginals $\PP(X_s)$. This is generally not true for graphs with cycles.
In many applications, however, the resulting loopy BP algorithm
exhibits excellent empirical performance~\citep{MurWeiJor99}. Several theoretical studies have also provided insight into the approximations made by loopy BP, partially justifying its application to graphs with cycles~\citep{WaiJor08,YedFreWei01}.

The learning problem in MRFs is to estimate the node and edge potentials, which is often done by maximizing the expected log-likelihood $\EE_{\bm{X}\sim\PP^\star(\bm{X})}[\log \PP(\bm{X})]$ of the model $\PP(\bm{X})$ with respect to the true distribution $\PP^\star(\bm{X})$. The resulting optimization problem usually requires solving a sequence of inference problems as an inner loop \citep{KolFri09}; BP is often deployed for this purpose. 
 
\vspace{-3mm}
\section{Properties of Belief Propagation}\label{sec:learnMRFviaBP}
\vspace{-3mm}

Our goal is to develop a nonparametric belief propagation algorithm, where the potentials are nonparametric functions learned from data, such that 
multimodal and other non-Gaussian statistical features can be captured. 
Most crucially, these potentials must be represented in  such a way that 
the message update in~\eq{eq:classicalBP} is computationally tractable. Before we go into the details of our kernel BP algorithm, we will first explain a key property of BP, which relates message updates to conditional expectations. When the messages are RKHS functions, these expectations can be evaluated efficiently.


\cite{YedFreWei01} showed  BP to be an iterative algorithm for minimizing the Bethe free energy, which is a variational approximation to the log-partition function, $\log Z$, in the MRF model $\PP(\bm{X})$. The beliefs are fixed points of BP algorithm if and only if they are zero gradient points of the Bethe free energy.
In Section 5 of the Appendix, we show maximum likelihood learning of MRFs using BP results in the following equality, which  relates 
the conditional of the true distribution, the learned potentials, and the fixed point messages,
\begin{align}
  \PP^\star(X_t|X_s) = \frac{\Psi_{st}(X_s,X_t)\Psi_t(X_t) \prod_{u\setminus s} m_{ut}^\star(X_t)}{m_{ts}^\star(X_s)}, \label{eq:momentMatching}
\end{align} 
where $\PP^\star(X_s)$ and $m_{ts}^\star(X_s)$ are assumed strictly positive.
\citet[Section 4]{WaiJaaWill03b} derived a similar relation, but for discrete variables under the exponential family setting. By contrast, we do not assume an exponential family model,  and our reasoning applies to continuous variables. 
A further distinction is that Wainwright et al. specify the node potential  $\Psi_{s}(X_s)=\PP^\star(X_s)$ and edge potential $\Psi(X_{s},X_{t})=\PP^\star(X_{s},X_{t})\PP^\star(X_{s})^{-1}\PP^\star(X_{t})^{-1}$, which represent just one possible choice among many that satisfies~\eq{eq:momentMatching}. 
Indeed, we next  show that in order to run BP for subsequent inference, we do not need to commit to a particular choice for $\Psi_{s}(X_s)$ and $\Psi(X_{s},X_{t})$, nor do we need to optimize to learn $\Psi_{s}(X_s)$ and $\Psi(X_{s},X_{t})$. 

We start by dividing both sides of~\eq{eq:classicalBP} by $m_{ts}^\star(X_s)$, and introducing $1=\prod_{u\setminus s} \frac{m_{ut}^\star(X_t)}{m_{ut}^\star(X_t)}$,
\begin{align}
  \frac{m_{ts}(X_{s})}{m_{ts}^\star(X_s)} = \int_{\Xcal} &~\frac{\Psi_{st}(X_s,X_t)\Psi_t(X_t) \prod_{u\setminus s} m_{ut}^\star(X_t)}{m_{ts}^\star(X_s)} \nonumber \\
  & \times \prod\nolimits_{u\setminus s} \frac{m_{ut}(X_{t})}{m_{ut}^\star(X_t)}~dX_t.
  \label{eq:divideclassicalBP}    
\end{align}
We next substitute the BP fixed point relation~\eq{eq:momentMatching} into \eq{eq:divideclassicalBP}, and reparametrize the messages $m_{ts}(X_{s})\leftarrow \frac{m_{ts}(X_{s})}{m_{ts}^\star(X_s)}$, to obtain the following property for BP updates (see Section 6 in the Appendix for details):
\vspace{-2.5mm}
\begin{property} \label{prop:condBPupdate}
If we learn an MRF using BP and subsequently use the learned potentials for inference, 
BP updates can be viewed as conditional expectations,
\begin{align}
  m_{ts}(X_{s})  
  & = \int_{\Xcal}\PP^\star(X_{t}|X_{s})\prod\nolimits_{u\setminus s}m_{ut}(X_{t})~dX_{t} \nonumber \\
  & = 
  \EE_{X_{t}|X_{s}}\left[\prod\nolimits_{u\setminus s}m_{ut}(X_{t})\right].\label{eq:condBPupdate}  
\end{align}
\vspace{-5mm}
\end{property}
Using similar reasoning, the node beliefs on convergence of BP take the form $\BB(X_s) \propto \PP^\star(X_{s})\prod\nolimits_{t\in\Gamma_{s}}m_{ts}^\star(X_{s})$. 
In the absence of external evidence, a fixed point occurs at the true node marginals,~i.e.,~$\BB(X_s)\propto \PP^\star(X_s)$ for all $s\in\Vcal$. 
Typically there can be many evidence variables, and the belief is then an estimate of the true conditional distribution given the evidence.

The above property of BP immediately suggests that if belief propagation is the inference algorithm of choice, then MRFs can be learned very simply: given training data drawn from $\PP^\star(\bm{X})$, the empirical conditionals $\widehat{\PP}(X_t|X_s)$ are estimated (either in parametric form, or nonparametrically), and the conditional expectations are evaluated using these estimates.
Evidence can also be incorporated straightforwardly: if an observation $x_{t}$ is made at node $t$, the message from $t$ to its neighbor $s$ is simply the empirical likelihood function $m_{ts}(X_{s})\propto\widehat{\PP}(x_{t}|X_{s})$, where we  use lowercase to denote observed variables with fixed values, and capitalize unobserved random variables.

With respect to kernel belief propagation, our key insight from Property~\ref{prop:condBPupdate}, however, is that we need not explicitly recover the empirical conditionals $\widehat{\PP}(X_t|X_s)$ as an intermediate step, as long as we can compute the conditional expectation directly. We will pursue this approach next.

\vspace{-3mm}  
\section{Kernel Belief Propagation\label{sec:Locally-consistent-belief}}
\vspace{-3mm}  

We develop a novel kernelization of belief
propagation, based on Hilbert space embeddings of conditional
distributions~\citep{SonHuaSmoFuk09}, which generalizes an earlier
kernel algorithm for exact inference on trees~\citep{SonGreGue10}. 
As might be expected, the kernel implementation of
the BP updates in  (\ref{eq:condBPupdate}) 
is nearly identical to 
the earlier tree
algorithm, the main difference being that we now consider graphs with loops, and iterate 
until convergence (rather than obtaining an exact solution in a single pass). 
This difference turns out to have major implications for the implementation:
 the earlier solution of Song et al. is polynomial
in the sample size, which was not an issue for the the smaller trees considered by
Song et al., but becomes expensive for the large, loopy graphical models 
we address in our experiments. We defer the issue of efficient implementation
to  Section \ref{sec:Efficient-kernel-message}, where we present a novel 
approximation strategy for kernel BP  which achieves constant time message updates.

In the present section, we will provide a detailed derivation of kernel BP 
in accordance with~\citet{SonGreGue10}. 
While the immediate purpose is to make 
the paper self-contained, there are two further important reasons:
 to provide
the background  necessary in understanding our efficient kernel BP updates 
in Section \ref{sec:Efficient-kernel-message}, and 
to demonstrate how kernel BP differs from the competing Gaussian mixture and particle based BP approaches in Section \ref{sec:GMandPBP} 
(which was not addressed in earlier work on kernel tree graphical models).



\vspace{-3mm}  
\subsection{Message Representations}
\vspace{-3mm}  


We begin with a description of the properties of a message $m_{ut}(x_{t})$,
given it is in the reproducing kernel Hilbert space (RKHS) $\Fcal$ of functions on the separable
metric space $\mathcal{X}$ \citep{Aronszajn50,SchSmo02}.
As we will see, the advantage of this assumption is that 
the update procedure can be expressed as a linear operation in the RKHS,
and results in new messages
 that are likewise RKHS functions. 
The RKHS $\Fcal$ is defined in terms of a unique positive definite kernel
$k(x_s,x_s')$ with the reproducing property 
$
  \inner{m_{ts}(\cdot)}{k(x_s, \cdot)}_{\Fcal}  = m_{ts}(x_s),
$
where $k(x_s, \cdot)$ indicates that one argument of the kernel is fixed at $x_s$.
Thus, we can view the evaluation of message $m_{ts}$ at any point $x_s\in\Xcal$ as 
a linear operation in $\Fcal$: we call $k(x_s, \cdot)$ the {\em representer
of evaluation} at $x_s$, and use the shorthand $k(x_s, \cdot)=\phi(x_s)$. Note that $k(x_s,x_s')=\inner{\phi(x_s)}{\phi(x_s')}_{\Fcal}$;
the kernel encodes the degree of similarity between $x_s$ and $x'_s$.
The restriction of messages to RKHS functions need not be onerous: on compact domains, universal kernels \citep[in the sense of][]{Steinwart01b} are dense in the space
of bounded continuous functions (e.g.,~the Gaussian RBF kernel $k(x_s,x_s') = \exp(-\sigma \nbr{x_s-x_s'}^2)$ is universal).
Kernels may be defined when dealing with random variables on additional
domains, such as strings, graphs, and groups. 


\vspace{-3mm}
\subsection{Kernel BP Message Updates}
\vspace{-3mm}

We next define a representation for message updates, under the assumption that  messages are RKHS functions. 
For simplicity, we first establish a result for a  three node chain, where the middle node $t$ incorporates an incoming message from $u$, and then generates an outgoing message to $s$ (we will deal with multiple incoming messages later). 
In this case, the outgoing message $m_{ts}(x_s)$ evaluated at $x_s$ simplifies to $m_{ts}(x_s) = \EE_{X_t|x_s}[m_{ut}(X_t)]$.  
 Under some regularity conditions for the integral, we can rewrite message updates as inner products, $m_{ts}(x_s) = \EE_{X_t|x_s}[\inner{m_{ut}}{\phi(X_t)}_{\Fcal}]= \inner{m_{ut}}{\EE_{X_t|x_s}[\phi(X_t)]}_{\Fcal}$ using the reproducing property of the RKHS. 
We refer to $\mu_{X_t|x_s}:=\EE_{X_t|x_s}[\phi(X_t)]\in \Fcal$ as the feature space embedding of the conditional distribution $\PP(X_t|x_s)$. If we can estimate this quantity directly from data,  we can perform message updates via a simple inner product, 
avoiding a two-step procedure where the conditional distribution is first estimated and the expectation then taken.

An expression for the conditional distribution embedding was proposed by \cite{SonHuaSmoFuk09}. 
We describe this expression by analogy with the conditioning operation for a Gaussian random vector $z\sim\Ncal(0,C)$, where we partition $z=(z_1^\top, z_2^\top)^\top$ such that $z_1\in\RR^d$ and $z_2\in\RR^{d'}$.  Given the covariances $C_{11}:=\EE[z_1 z_1^\top]$ and $C_{12}:=\EE[z_1 z_2^\top]$, we can write the conditional expectation $\EE[Z_1|z_2] =  C_{12}C_{22}^{-1}z_{2}$. We now generalize this notion to RKHSs. 
Following \cite{FukBacJor04}, we define the covariance operator $\Ccal_{X_sX_t}$ which allows us to compute the expectation of the product of function $f(X_s)$ and $g(X_t)$,~i.e.~$\EE_{X_sX_t}[f(X_s)g(X_t)]$, using linear operation in the RKHS. More formally, let $\Ccal_{X_sX_t}:\Fcal\mapsto\Fcal$ such that for all $f,g,h\in\Fcal$,
\begin{align}
  \EE_{X_sX_t}[f(X_s)g(X_t)]
  =&~
  \left\langle f,~\EE_{X_sX_t} \left[\phi(X_s)\otimes\phi(X_t)\right] g \right\rangle _{\Fcal} \nonumber \\
  =&~
  \left\langle f,~\Ccal_{X_sX_t}g\right\rangle _{\Fcal}
\end{align}
where we use tensor notation
$
(f\otimes g)h=f\left\langle g,h\right\rangle_{\Fcal}.
$
%
This can be understood by analogy with the finite dimensional case: if $x,y,z\in\RR^d$, then $(x\,y^\top) z = x (y^\top z)$; furthermore, $(x^\top x')(y^\top y')(z^\top z')=\inner{x\otimes y \otimes z}{~x'\otimes y' \otimes z'}_{\RR^{d^3}}$ given $x,y,z,x',y',z'\in\RR^d$. 
Based on covariance operators, Song et al. define a conditional embedding operator
which allow us to compute conditional expectations $\EE_{X_t|x_s}[f(X_t)]$ as linear operations in the RKHS. Let $\Ucal_{X_t|X_s}:= \Ccal_{X_tX_s}\Ccal_{X_sX_s}^{-1}$ such that for all $f\in\Fcal$, 
\begin{align}
  \EE_{X_t|x_s}[f(X_t)]
  =&~\inner{f}{~\EE_{X_t|x_s}[\phi(X_t)}_{\Fcal}
  =\inner{f}{~\mu_{X_t|x_s}}_{\Fcal} \nonumber \\
  =&~\inner{f}{~\Ucal_{X_t|X_s}\phi(x_s)}_{\Fcal}. \label{eq:condMean} 
\end{align}
Although we used the intuition from the Gaussian case in understanding this formula, it is important to note that the conditional embedding operator allows us to compute the conditional expectation of {\em any} $f\in\Fcal$, regardless of the distribution of the random variable in feature space (aside from the condition that $h(x_s):=\EE_{X_t|x_s}[f(X_t)]$ is in the RKHS on $x_s$, as noted by Song et al.). In particular, we do not need to assume the random variables have a Gaussian distribution in feature space (the definition of feature space Gaussian BP remains a challenging open problem: see Appendix, Section 7).

%

We can thus express the message update as a linear operation in the feature space,
\begin{align}
    m_{ts}(x_s)     
    = \inner{m_{ut}}{~\Ucal_{X_t|X_s}\phi(x_s)}_{\Fcal}.\nonumber
\end{align}
%
%
%
For multiple incoming messages, 
the message updates follow the same
reasoning as in the single message case, albeit with some additional notational complexity \citep[see also][]{SonGreGue10}.
We begin by defining a tensor product reproducing kernel Hilbert space
 $\Hcal:=\otimes^{d_t-1} \Fcal$, under which the product of incoming messages can be written as a single inner product. 
For a node $t$ with degree $d_t=|\Gamma_t|$, the product of incoming messages $m_{ut}$ from all neighbors except $s$ becomes
 an inner product in $\Hcal$, 
\begin{align}  
  \prod\nolimits_{u\setminus s} m_{ut}(X_t)
  &=\prod\nolimits_{u\setminus s} \inner{m_{ut}}{~\phi(X_t)}_{\Fcal} \nonumber \\
  &= \inner{\bigotimes\nolimits_{u\setminus s}m_{ut}}{~\xi(X_t)}_{\Hcal},
\end{align}
where $\xi(X_t):= \bigotimes\nolimits_{u\setminus s}\phi(X_t)$. 
The message update \eq{eq:condBPupdate} becomes
\begin{align}
\hspace{-3mm}
  m_{ts}(x_{s})=\inner{\bigotimes\nolimits_{u\setminus s}m_{ut}}{~\EE_{X_{t}|x_{s}}\sbr{\xi(X_t)}}_{\Hcal}.
\end{align}
By analogy with~\eq{eq:condMean}, we can define the conditional embedding operator
for the tensor product of features, such that
$\Ucal_{X_{t}^{\otimes}|X_{s}} : \Fcal\rightarrow\Fcal^{\otimes}$ satisfies
\begin{align}
  \hspace{-3mm}
  \mu_{X_t^\otimes|x_s}:=\EE_{X_{t}|x_{s}}\left[\xi(X_t)\right] = 
  \Ucal_{X_{t}^{\otimes}|x_{s}}\phi(x_{s}). 
\end{align}
As in the single variable case, $\Ucal_{X_{t}^{\otimes}|x_{s}}$ is defined in terms of a covariance operator $\Ccal_{X_t^\otimes X_s}:=\EE_{X_tX_s}[\xi(X_t)\otimes \phi(X_s)]$ in the tensor space, and the operator $\Ccal_{X_sX_s}$. 
The operator $\Ucal_{X_{t}^{\otimes}|X_{s}}$ takes the feature map $\phi(x_s)$ of the point on which we condition, and outputs the conditional expectation of the tensor product feature $\xi(X_t)$.
Consequently, we can express the message update as a linear operation, but in a tensor product feature space,
\begin{align}
  \hspace{-3mm}
  m_{ts}(x_{s})=
  \inner{\bigotimes\nolimits_{u\setminus s}m_{ut}}{~~\Ucal_{X_{t}^{\otimes}|X_{s}}\phi(x_{s})}_{\Hcal}.
  \label{eq:updateTensor}
\end{align}



The belief at a specific node $s$ can be computed as $\BB(X_s)=\PP^\star(X_s)\prod_{u\in\Gamma_{s}}m_{us}(X_{s})$ where the true marginal $\PP^\star(X_r)$ can be estimated using Parzen windows. If this is undesirable (for instance, on domains where density estimation cannot be performed), the belief can instead be expressed as a conditional embedding operator \citep{SonGreGue10}.

\vspace{-3mm}
\subsection{Learning Kernel Graphical Models}
\vspace{-3mm}

Given a training sample of $m$ pairs $\cbr{(x_t^i,x_s^i)}_{i=1}^m$ drawn~\iid~from $\PP^\star(X_t,X_s)$, we can represents messages and their updates based purely on these training examples. We first define feature matrices $\Phi=(\phi(x_t^1),\ldots,\phi(x_t^m))$, $\Upsilon=(\phi(x_s^1),\ldots,\phi(x_s^m))$ and $\Phi^\otimes=\rbr{\xi(x_t^1),\ldots,\xi(x_t^m)}$, 
and  corresponding kernel matrices $K=\Phi^\top \Phi$ and $L=\Upsilon^\top \Upsilon$. 
The assumption that messages are RKHS functions  means that messages can be represented as linear combinations of the training features $\Phi$,~i.e.,~$\widehat{m}_{ut}=\Phi \beta_{ut}$, where $\beta_{ut}\in\RR^m$. On this basis, \citet{SonHuaSmoFuk09} propose a direct regularized estimate of the conditional embedding operators
 from the data. This approach avoids explicit conditional density estimation, and directly provides the  tools needed for computing the RKHS message updates in~\eq{eq:updateTensor}. Following this approach, we first estimate the covariance operators $\widehat{\Ccal}_{X_tX_s}=\frac{1}{m}\Phi\Upsilon^\top$, $\widehat{\Ccal}_{X_t^\otimes X_s}=\frac{1}{m}\Phi^\otimes \Upsilon^\top$ and 
 $\widehat{\Ccal}_{X_sX_s}=\frac{1}{m}\Upsilon\Upsilon^\top$, and obtain an empirical estimate of the conditional embedding operator,
\begin{align}
  \hUcal_{X_t^\otimes|X_s} = \Phi^\otimes(L^\top+\lambda m I)^{-1}\Upsilon^\top, 
  \label{eq:empUpdateFullRank_prekernel}
\end{align}
where $\lambda$ is a regularization parameter. 
Note that we need not  compute the feature space covariance operators explicitly:  as we will see, all steps in kernel BP are carried out via operations on kernel matrices.

We now apply the empirical conditional embedding operator to obtain a finite sample message update for (\ref{eq:updateTensor}).
Since the incoming messages $\widehat{m}_{ut}$ can be expressed as $\widehat{m}_{ut}=\Phi \beta_{ut}$,
 the outgoing message $\widehat{m}_{ts}$ at $x_s$ is
\begin{align}
  &\inner{\bigotimes\nolimits_{u\setminus s}\Phi \beta_{ut}}{~~\Phi^\otimes(L+\lambda m I)^{-1}\Upsilon^\top\phi(x_s)}_{\Hcal} \nonumber \\
  =&\rbr{\bigodot\nolimits_{u\setminus s} K \beta_{ut}}^\top (L+\lambda m I)^{-1} \Upsilon^\top \phi(x_s)
  \label{eq:empUpdateFullRank}
\end{align}
where $\bigodot$ is the elementwise vector product. If we define $\beta_{ts} = (L+\lambda m I)^{-1} (\bigodot\nolimits_{u\setminus s} K \beta_{ut})$, then the outgoing message can be expressed as $\widehat{m}_{ts}=\Upsilon \beta_{ts}$. 
In other words, given incoming messages expressed as linear combinations of feature mapped training samples from $X_t$, the outgoing message will likewise be a weighted linear combination of feature mapped training samples from $X_s$. 
Importantly,  only $m$ mapped points will be  used
to express the outgoing message, regardless of the number of incoming messages
or the number of points used to express each incoming message. Thus the complexity of message representation does not increase with BP iterations or degree of a node. 

Although we have identified the model parameters with specific edges $(s,t)$, our approach extends straightforwardly to a templatized model, where parameters are shared across multiple edges (this setting is often natural in image processing, for instance). Empirical estimates of the parameters are  computed on the pooled observations.

The computational complexity of the finite sample BP update in (\ref{eq:empUpdateFullRank}) is polynomial in term of the number of training samples. 
Assuming a preprocessing step of cost $O(m^{3})$ to compute the  matrix inverses, the update for a \emph{single} message costs $O(m^{2}d_{\mathrm{max}})$ where $d_{\mathrm{max}}$ is the maximum degree of a node in the MRF. 
While this  is reasonable in comparison with competing nonparametric approaches (see  Section \ref{sec:GMandPBP} and the experiments), 
and works well for smaller graphs and trees,
a polynomial time update can  be costly for very large $m$, and for graphical models with loops (where many iterations of the message updates are needed).
 In  Section \ref{sec:Efficient-kernel-message}, we  develop a message approximation strategy which reduces this cost substantially.

\vspace{-3mm}
\section{Constant Time Message Updates\label{sec:Efficient-kernel-message}}
\vspace{-3mm}

In this section, we formulate a more computationally efficient alternative to the full rank update
in (\ref{eq:empUpdateFullRank}). Our goal
is to limit the computational cost of each update to 
$O(\ell^{2}d_{\mathrm{max}})$
where $\ell\ll m$. We will require a one-off preprocessing step which is linear in $m$. This efficient message passing procedure makes  kernel BP practical even for very large graphical models and/or training set sizes. 

\vspace{-3mm}
\subsection{Approximating Feature Matrices}\label{sec:approxFeatMat}
\vspace{-3mm}


The key idea of the preprocessing step is to approximate messages in the RKHS with a few informative basis functions, and to estimate these basis functions in a data dependent way. This is achieved by approximating the feature matrix $\Phi$ as a weighted combination of a subset of its columns. That is, $\Phi\approx \Phi_{\Ical}W_t$, where $\Ical:=\cbr{i_1,\ldots,i_{\ell}}\subseteq \cbr{1,\ldots,m}$, $W_t$ has dimension $\ell\times m$, and $\Phi_{\Ical}=(\phi(x_t^{i_1}),\ldots,\phi(x_t^{i_\ell}))$ is a submatrix formed by taking the columns of $\Phi$ corresponding to the indices in $\Ical$.  Likewise, we approximate $\Upsilon\approx \Upsilon_{\Jcal} W_s$, assuming  $|\Jcal|=\ell$ for simplicity. We thus can approximate the kernel matrices as low rank factorizations,~i.e.,~$K\approx W_t^\top K_{\Ical\Ical} W_t$ and $L=W_s^\top L_{\Jcal\Jcal} W_s$, where $K_{\Ical\Ical}:=\Phi_{\Ical}^\top\Phi_{\Ical}$ and $L_{\Jcal\Jcal}=\Upsilon_{\Jcal}^\top\Upsilon_{\Jcal}$. 

A common way to obtain the approximation $\Phi\approx \Phi_{\Ical}W_t$ is via a Gram-Schmidt orthogonalization procedure in feature space, where an incomplete set of $\ell$ orthonormal basis vectors $Q:=(q_t^1,\ldots,q_t^{\ell})$ is constructed from a greedily selected subset of the data, chosen to minimize the reconstruction error~\cite[p.126]{ShaCri04}. 
The original feature matrix can be approximately expressed using this basis subset as $\Phi\approx Q R$ where $R\in\RR^{\ell\times m}$ are the coefficients under the new basis. There is a simple relation between  $Q$ and the chosen data points $\Phi_{\Ical}$,~i.e.,~ $Q=\Phi_{\Ical}R_{\Ical}^{-1}$, where $R_{\Ical}$ is the submatrix formed by taking the columns of $R$ corresponding to $\Ical$. It follows that
 $W_t = R_{\Ical}^{-1} R$. 
All operations involved in Gram-Schmidt orthogonalization are linear in feature space, and the entries of $R$ can  be computed based solely on kernel values $k(x_t,x_t')$. The cost of performing this orthogonalization is $O(m\ell^{2})$. 
The number $\ell$ of chosen basis vectors is inversely related to the approximation error or residual 
$\epsilon=\max_{i} \|\phi(x_t^i) - \Phi_{\Ical} W_t^i \|_{\Fcal}$ 
($W_t^i$ denotes column $i$ of $W_t$). 
In many cases of interest  (for instance, when a Gaussian RBF kernel is used), a small $\ell\ll m$ will be sufficient to obtain a small residual $\epsilon$ for the feature matrix, 
due to the fast decay of the eigenspectrum in feature space~\cite[Appendix C]{BacJor02}. 

%

\vspace{-3mm}
\subsection{Approximating Tensor Features}
\vspace{-3mm}

The approximations $\Phi\approx\Phi_{\Ical}W_t$ and $\Upsilon\approx\Upsilon_{J}W_s$, and associated low rank kernel approximations 
 are insufficient for a constant time approximate algorithm, however. 
In fact, directly applying these results will only lead to a linear time approximate algorithm: this can be seen by  replacing the kernel matrices in~\eq{eq:empUpdateFullRank} by their low rank approximations. 

To achieve a constant approximate update, our strategy is to go a step further: in addition to approximating the kernel matrices, we further approximate the tensor product feature matrix in equation~\eq{eq:empUpdateFullRank_prekernel}, $\Phi^{\otimes}\approx \Phi_{\Ical'}^\otimes W_t^\otimes$ ($W_t^\otimes\in\RR^{\ell'\times m}$). 
Crucially, the individual kernel matrix approximations neglect to account for  the subsequent tensor product of these messages. 
By contrast, our proposed approach also approximates the tensor product directly.
The computational advantage of a direct tensor approximation approach is substantial in practice (a comparison between exact kernel BP and its constant and linear time approximations can be found in Section 3 of the Appendix) .



The decomposition procedure for tensor $\Phi^{\otimes}\approx \Phi_{\Ical'}^\otimes W_t^\otimes$ follows exactly the same steps as in the original feature space, but using the kernel $k^{d_t-1}(x_t,x_t')$, and yielding an incomplete orthonormal basis in the tensor product space. In general the index sets $\Ical'\neq\Ical$, meaning they select different training points to construct the basis functions. Furthermore, the size $\ell'$ of $\Ical'$ is not equal to the size $\ell$ of $\Ical$ for a given approximation error $\epsilon$. Typically $\ell'>\ell$, since the tensor product space has a slower decaying spectrum, however we will write $\ell$ 
in place of $\ell'$ to simplify notation.


\vspace{-3mm}
\subsection{Constant Time Approximate Updates}
\vspace{-3mm}

We now compute the message updates based on the various low rank approximations.
The incoming messages and the conditional embedding operators become
\begin{align}   
  \bigotimes\nolimits_{u\setminus s} m_{ut}
  &\approx 
  \bigotimes\nolimits_{u\setminus s}\Phi_{\Ical} W_t \beta_{ut},\\
  \widetilde{\Ucal}_{X_t^\otimes|X_s}\phi(x_s)
  &\approx
  \Phi_{\Ical'}^\otimes W_{ts} \Upsilon_{\Jcal}^\top \phi(x_s), \label{eq:empUpdateLowRank_prekernel}
\end{align}
where $W_{ts}:=W_t^\otimes (W_s^\top L_{\Jcal\Jcal} W_s+\lambda m I)^{-1}W_s^\top$. If we reparametrize the messages $m_{ut}$ as $m_{ut}=\Phi_{\Ical} \alpha_{ut}$ where  $\alpha_{ut}:=W_t\beta_{ut}$, we can express the message updates for $m_{ts}(x_s)$ as
\begin{align}
  m_{ts}(x_s)\approx
  \rbr{\bigodot\nolimits_{u\setminus s} K_{\Ical'\Ical} \alpha_{ut}}^\top W_{st} \Upsilon_{\Jcal}^\top \phi(x_s),
  \label{eq:approxBP}
\end{align}
where $K_{\Ical'\Ical}$  denotes the submatrix of  $K$ with rows indexed $\Ical'$ and columns indexed $\Ical$. The outgoing message $m_{ts}$ can also be reparametrized as a vector $\alpha_{ts}= W_{st}^\top \rbr{\bigodot\nolimits_{u\setminus s} K_{\Ical'\Ical} \alpha_{ut}}$. 
In short, the message from $t$ to $s$ is a weighted linear combination
of the $\ell$ vectors in $\Upsilon_{\Jcal}$. 

We note that $W_{ts}$ can be computed efficiently prior to the message update step, since
 $W_t^\otimes (W_s^\top L_{\Jcal\Jcal} W_s+\lambda m I)^{-1}W_s^\top = W_t^\otimes W_s^\top (W_sW_s^\top + \lambda m L_{\Jcal\Jcal}^{-1})^{-1} L_{\Jcal\Jcal}^{-1}$ via the Woodbury expansion of the matrix inverse. 
In the latter form, matrix products $W_sW_s^\top$ and $W_t^\otimes W_s^\top$ cost $O(\ell^{2}m)$; the remaining operations (size $\ell$ matrix products and
inversions) are  significantly less costly at $O(\ell^{3})$. This initialization cost of $O(\ell^3 + \ell^2 m)$ need only be borne once. 

The cost of updating a single message
$m_{ts}$ in~\eq{eq:approxBP}~becomes $O(\ell^2d_{\mathrm{max}})$ where $d_{\rm{max}}$ is the maximum degree of a node. This also means that our approximate message update scheme will be independent of the number of training examples. With these approximate messages, the evaluation of the belief $\widehat{\BB}(x_r)$ of a node $r$ at $x_r$ can be carried out in time $O(\ell d_{\max})$. 

Finally, approximating the tensor features introduces additional error into each message update. This is caused by the difference between the full rank conditional embedding operator $\hUcal_{X_t^\otimes|X_s}$ in~\eq{eq:empUpdateFullRank_prekernel} and its low rank counterpart $\widetilde{\Ucal}_{X_t^\otimes|X_s}$ in~\eq{eq:empUpdateLowRank_prekernel}. Under suitable conditions, this difference is bounded by the feature approximation error $\epsilon$,~i.e.,~$\|\hUcal_{X_t^\otimes|X_s} - \widetilde{\Ucal}_{X_t^\otimes|X_s}\|_{HS} \leq 2\epsilon(\lambda^{-1} + \lambda^{-3/2})$ (see Section 8 of the Appendix for details). 



\vspace{-3mm}
\section{Gaussian Mixture And Particle BP}\label{sec:GMandPBP}
\vspace{-3mm}



We briefly review two state-of-the-art approaches to nonparametric
belief propagation: Gaussian Mixture BP \citep{SudIhlFreWil03} and
Particle BP \citep{IhlMcA09}.  By contrast with our approach, we must
provide these algorithms in advance with an estimate of the
conditional density $\PP^\star(X_t|X_s)$, to compute the conditional
expectation in (\ref{eq:condBPupdate}). For Gaussian
Mixture BP, this conditional density must take the form of a mixture of Gaussians.
We describe how we learn the conditional density from data, and then
show how the two algorithms use it for inference.

A direct approach to estimating the conditional density $\PP^\star(X_t|X_s)$
would be to take the ratio of the joint empirical density
 to the marginal empirical density.
The ratio
of mixtures of Gaussians is not itself a mixture of Gaussians,
however, so this approach is not suitable for Gaussian Mixture BP
(indeed, message updates using this ratio of mixtures would be
non-trivial, and we are not aware of any such inference approach).
  We propose instead to learn $\PP^\star(X_t|X_s)$ directly from training data
following \citet{SugTakSuzKanetal10}, who provide an
estimate in the form of a mixture of Gaussians (see Section 1 of the Appendix for details). We emphasize that the updates bear no
resemblance to our kernel updates in (\ref{eq:empUpdateFullRank}),
which do not attempt density ratio estimation.

Given the estimated $\widehat{\PP}(X_t|X_s)$ as input, each nonparametric
inference method takes a different approach.  Gaussian
mixture BP assumes incoming messages to be a mixture of $b$
Gaussians. The product of $d_t$ incoming messages to node $t$ then
contains $b^{d_{t}}$ Gaussians. This exponential blow-up is avoided by
replacing the exact update with an approximation. An overview of approximation approaches can be found in \cite{FaultDet}; we used an efficient KD-tree method of  \cite{KD-tree} for performing the approximation step.
Particle BP represents the incoming messages using a common
set of particles.
These particles must 
 be re-drawn via Metropolis-Hastings at each
node and BP iteration, which is  costly 
(although in practice, it is sufficient to resample periodically, rather than strictly at every iteration).  
By contrast, our updates are simply matrix-vector products.
See Appendix for further discussion.

\vspace{-3mm}
\section{Experiments}\label{sec:experiments}
\vspace{-3mm}

We performed four sets of experiments. The first two were image
denoising and depth prediction problems, where we show that kernel BP is
superior to discrete, Gaussian mixture and particle BP in both speed and accuracy, using a GraphLab implementation of each~\citep{LowGonKyrBicetal10}. The remaining two experiments were protein structure and paper category prediction problems,
where domain-specific kernels were crucial
(for the latter see Appendix, Sec. 4).

\setlength{\tabcolsep}{2pt}
\begin{figure}
  \centering 
  \begin{tabular}{cc}
    \includegraphics[width=0.20\columnwidth]{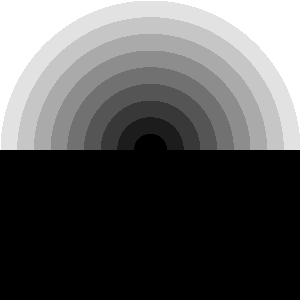} &
    \includegraphics[width=0.20\columnwidth]{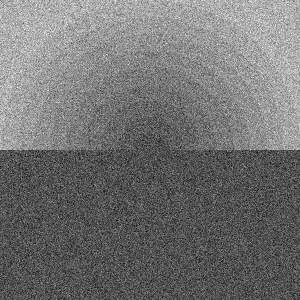} \\
    (a) {\footnotesize Sunset image} & (b) {\footnotesize Noisy image} \\    
   \includegraphics[width=0.40\columnwidth]{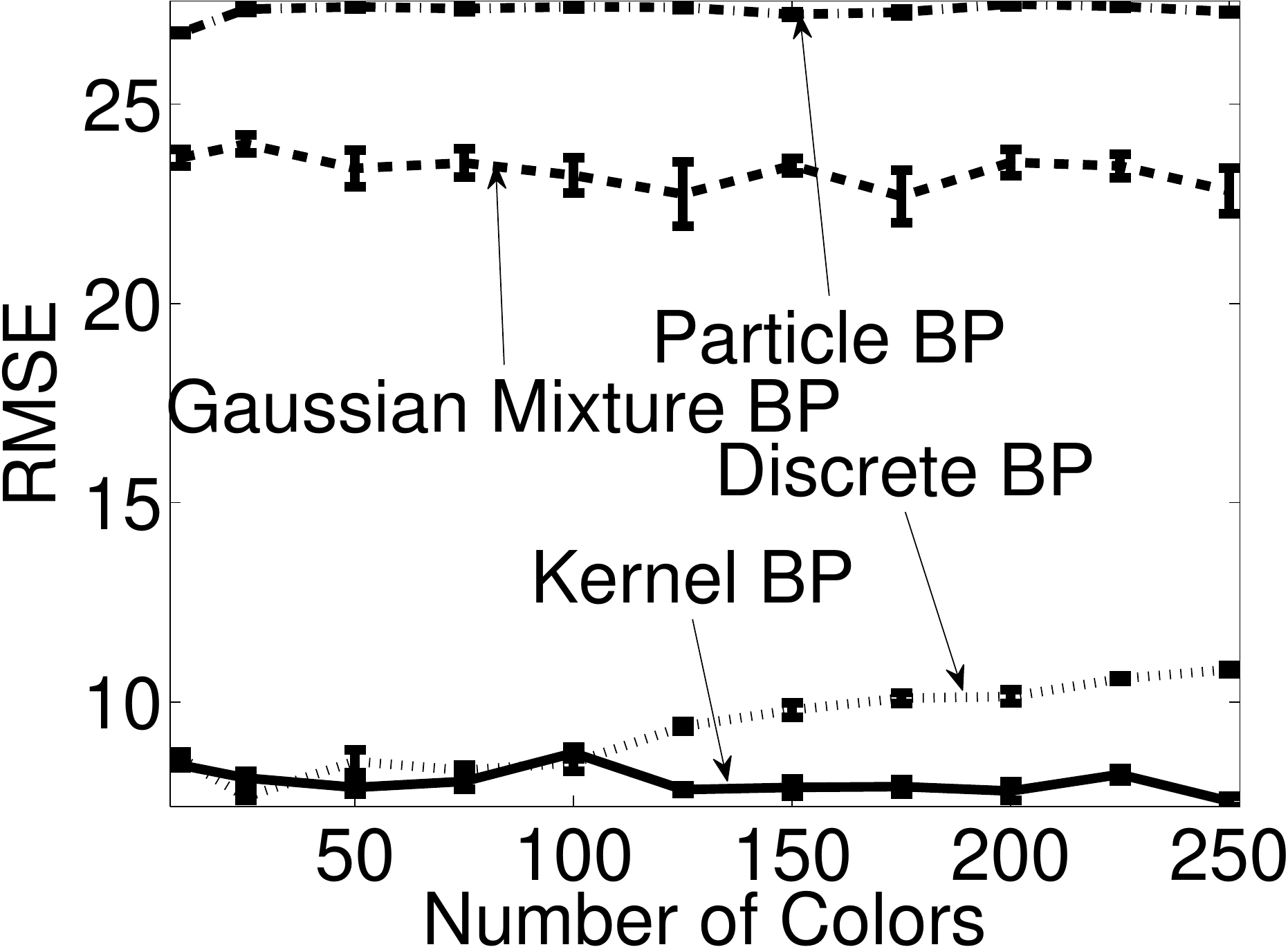} & 
   \includegraphics[width=0.40\columnwidth]{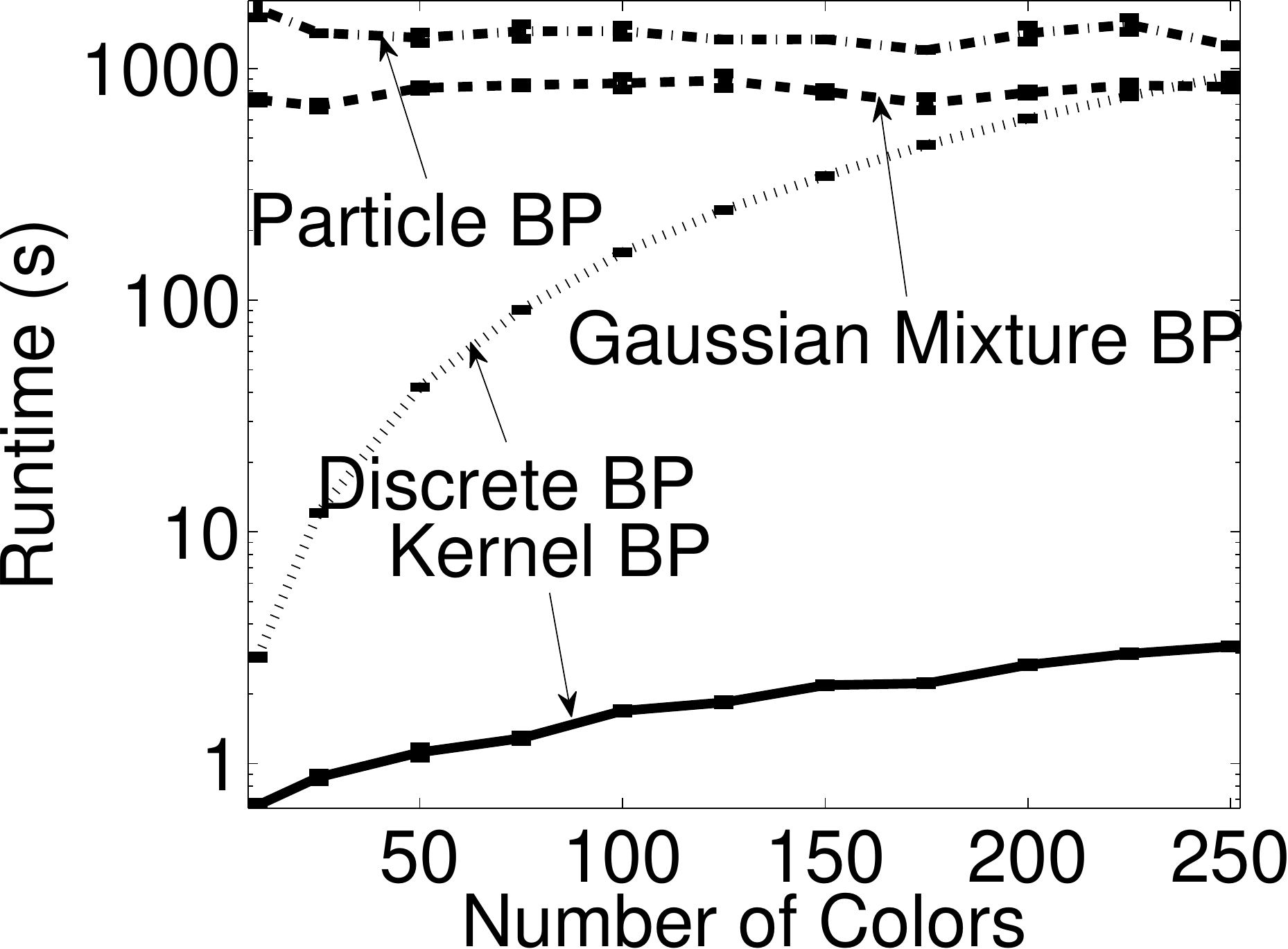} \\    
    (c) {\footnotesize Denoising error} & (d) {\footnotesize Runtime}
    \vspace{-3mm}
  \end{tabular}
  \caption{\footnotesize Average denoising error and runtime of kernel BP compared to discrete, Gaussian mixture and particle BP over 10 test images with varying numbers of rings. Runtimes are plotted on a logarithmic scale.}
  \label{fig:exp1}
  \vspace{-7mm}
\end{figure}  

{\bf Image denoising:} In our first experiment, the data consisted of grayscale images of size $100\times 100$, resembling a sunset (Figure~\ref{fig:exp1}(a)). 
The number of colors (gray levels) in the images ranged across 10, 25, 50, 75, 100, 125, 150, 175, 200, 225 and 250, 
with gray levels varying evenly from 0 to 255 from the innermost ring of the sunset to the outermost. As we increased the number of colors, the grayscale transition became increasingly smooth.   
Our goal was to recover the original images from noisy versions, to which we had added zero mean Gaussian noise with $\sigma=30$.
We compared the denoising performance and runtimes of discrete, Gaussian mixture, particle, and kernel BP. 

The topology of our graphical model was a grid of hidden noise-free pixels with noisy observations made at each. The maximum degree of a node was 5 (four neighbours and an observation), and we used a template model where both the edge potentials and the likelihood functions were shared across all variables. We generated a pair of noise-free and noisy images as training data, at each color number. For kernel BP, we learned both the likelihood function and the embedding operators nonparametrically from the data. We used a Gaussian RBF kernel $k(x,x')$, with kernel bandwidth set at the median distance between training points, and residual $\epsilon=10^{-3}$ as the stopping criterion for the feature approximation (see definition of $\epsilon$ in Section \ref{sec:approxFeatMat}). For discrete, Gaussian mixture, and particle BP, we learned the edge potentials from data, but supplied the {\em true} likelihood of the observation given the hidden pixel (i.e.,~a Gaussian with standard deviation 30). 
This gave competing methods an important {\em a priori} advantage over kernel BP: in spite of this, kernel BP still outperformed competing approaches in speed and accuracy.

In Figure~\ref{fig:exp1}(c) and (d), we report the average denoising performance (RMSE: root mean square error) and runtime over 30 BP iterations, using 10 independently generated noisy test images. The RMSE of kernel BP is significantly lower than Gaussian mixture and particle BP  for all numbers of colors. Although the RMSE of discrete BP is about the same as kernel BP when the number of  colors is small, its performance becomes worse than kernel BP as the number of colors increases beyond 100 (despite discrete BP receiving the true observation likelihood in advance).  In terms of speed,  kernel BP has a considerable advantage over the alternatives: the runtime of KBP is barely affected by the number of colors.  
For discrete BP, the scaling is approximately square in the number of colors. For Gaussian mixture and particle BP, the runtimes are orders of magnitude longer than kernel BP, and are affected by the variability of the resampling algorithm. 

\begin{figure}
  \centering
  \begin{tabular}{cc}
 \includegraphics[width=0.21\columnwidth]{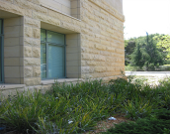}
    \includegraphics[width=0.21\columnwidth]{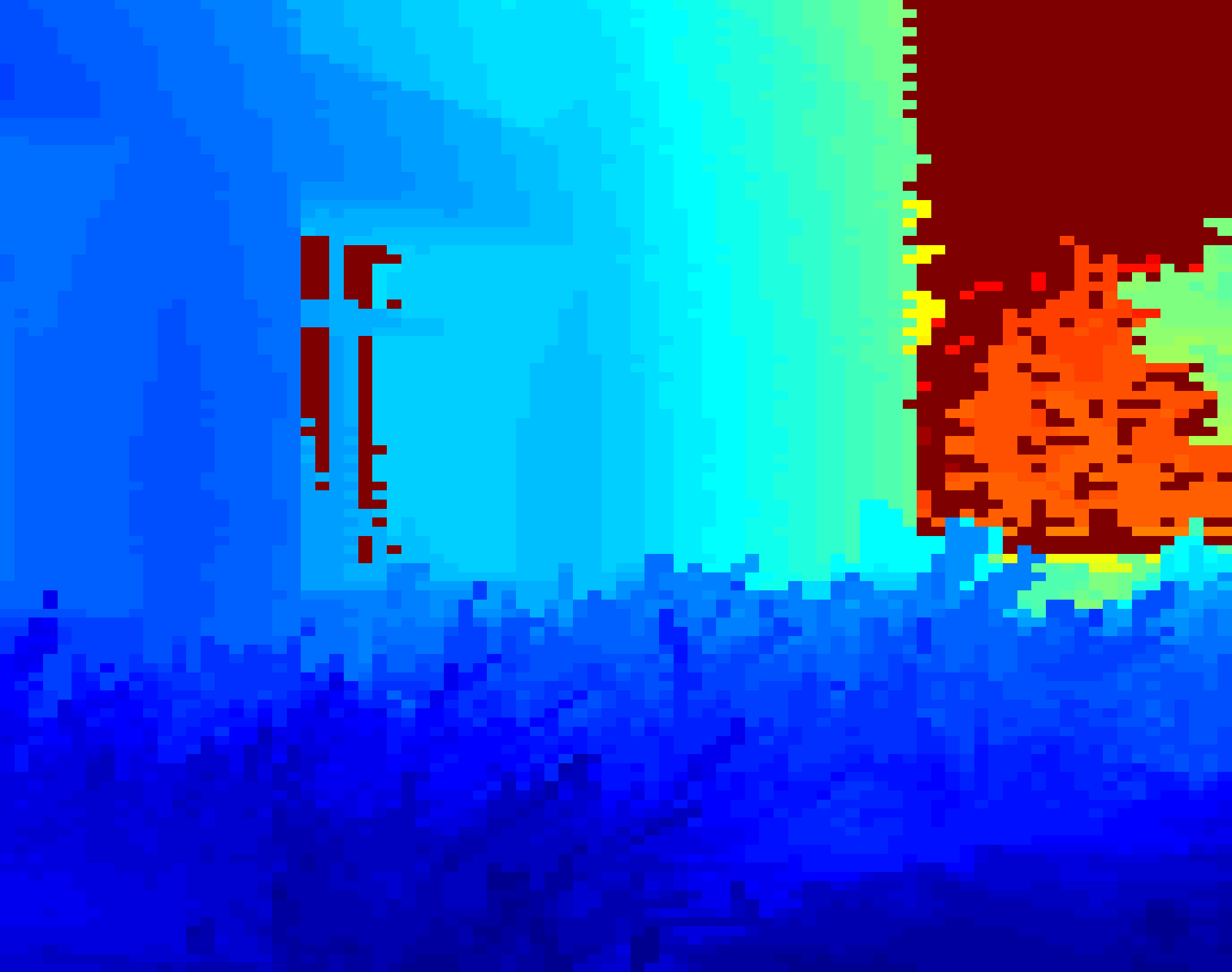} &
    \includegraphics[width=0.40\columnwidth]{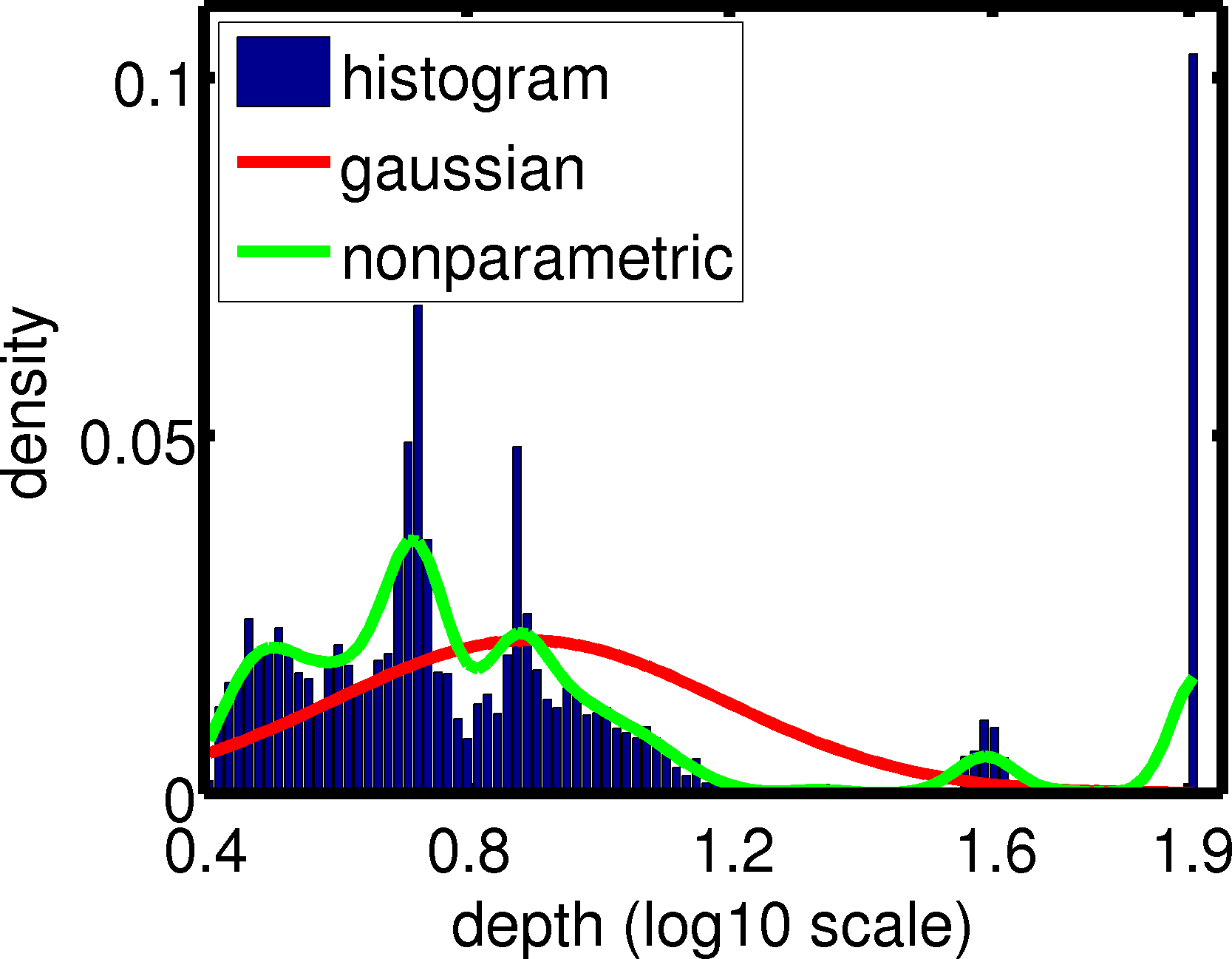} \\    
    (a) {\footnotesize Image and depth pair} & (b) {\footnotesize Depth distribution} \\
    \includegraphics[width=0.40\columnwidth]{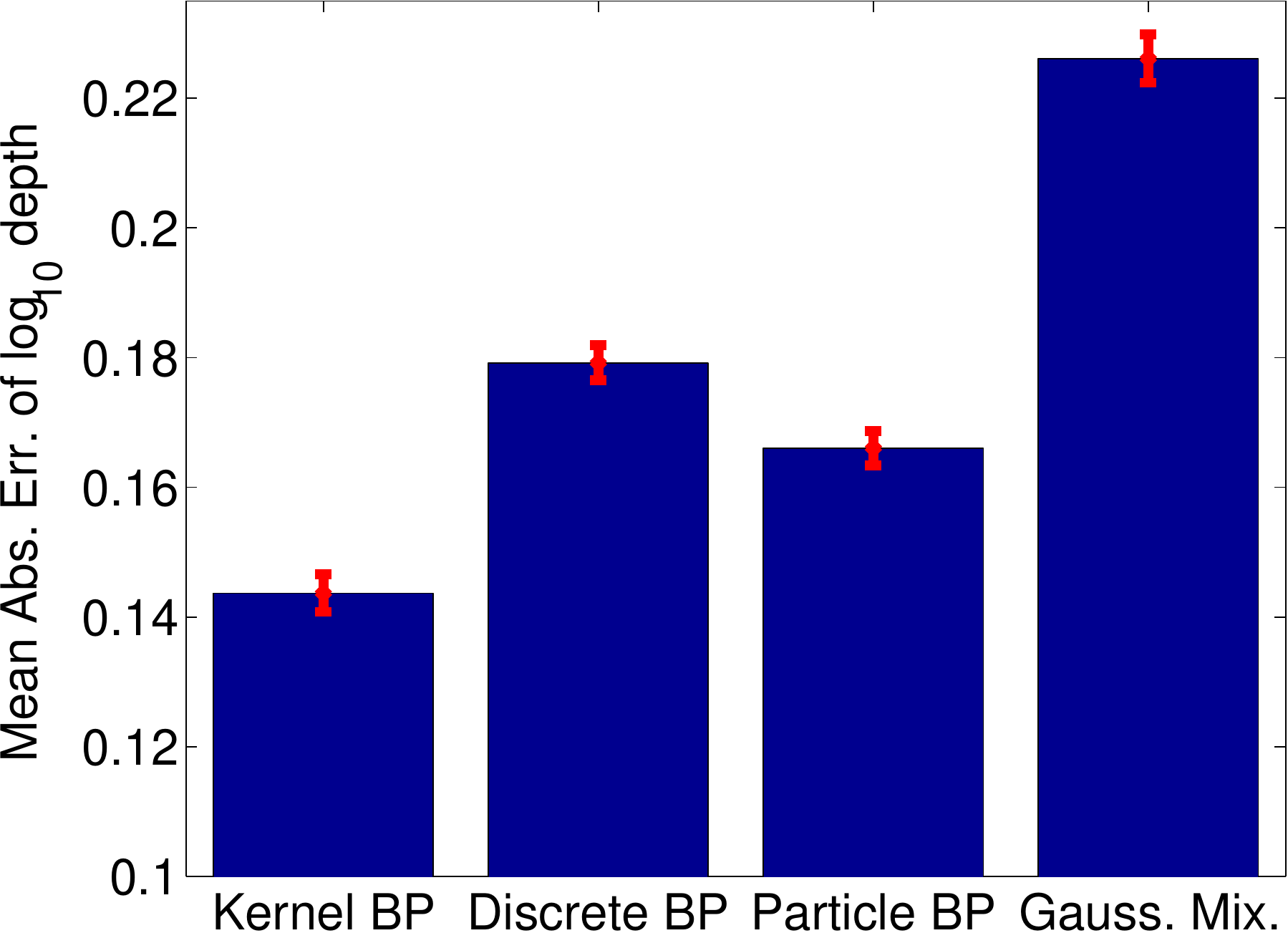} &    
    \includegraphics[width=0.40\columnwidth]{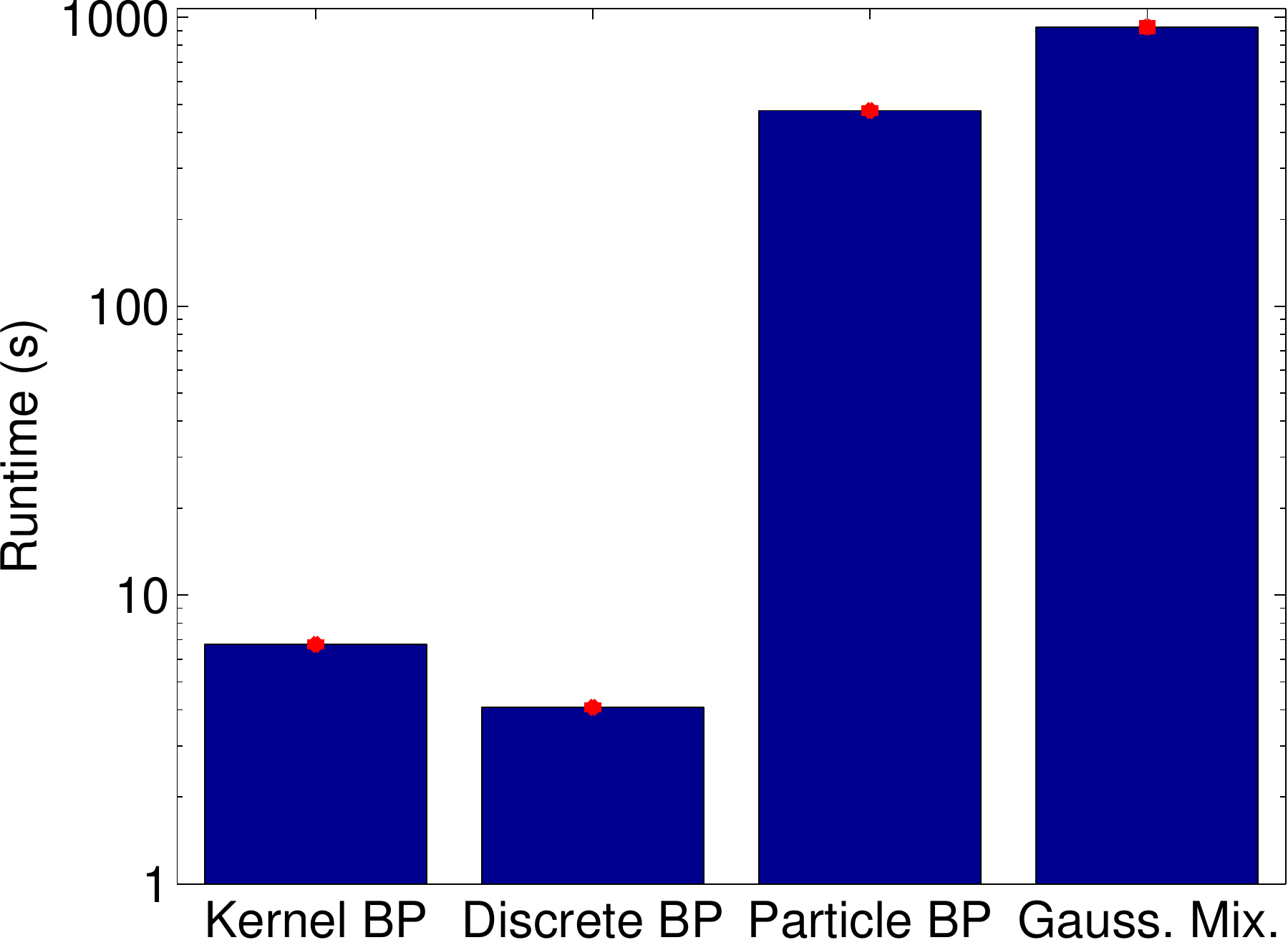} \\
    (c) {\footnotesize Depth prediction error} & (d) {\footnotesize Runtime}
    \vspace{-3mm}
  \end{tabular}
  \caption{\footnotesize Average depth prediction error and runtime of kernel BP compared to discrete, Gaussian mixture and particle BP over 274 images. Runtimes are on a logarithmic scale.
  }   
  \label{fig:depth_results}    
  \vspace{-5mm}  
\end{figure}

{\bf Predicting depth from 2D images:} The prediction of  3D depth information from 2D image features is a difficult inference problem, as the depth may be  ambiguous: similar features can occur at different depths. This creates a multimodal depth distribution  given the image feature. Furthermore, the marginal distribution of the depth can itself be multimodal, which makes the Gaussian approximation a poor choice~(see Figure~\ref{fig:depth_results}(b)). To make a spatially consistent prediction of the depth map, we  formulated the problem as an undirected graphical model, where a depth variable $y_i\in\RR$ was associated with each patch of an image, and these variables were connected according to a 2D grid topology. Each  hidden depth variable was linked to an image feature variable $x_i\in\RR^{273}$ for the corresponding patch. 
This formulation resulted in a graphical model with $9,202=107\times 86$ continuous depth variables, and a maximum node degree of 5. 
Due to the way the images were taken (upright), we used a templatized model where horizontal edges in a row shared the same potential, vertical edges at the same height shared the same potential, and patches at the same row shared the same likelihood function. Both the edge potentials between adjacent depth variables and the likelihood function between image feature and depth were unknown, and were learned from data.


We used a set of 274 images taken on the Stanford campus, including both indoor and outdoor scenes~\citep{SaxSunNg09}. Images were divided into patches of size 107 by 86, with the corresponding depth map for each patch obtained using 3D laser scanners (\eg,~Figure~\ref{fig:depth_results}(a)). Each patch was represented by a 273 dimensional feature vector, which contained both local features (such as color and texture) and relative features (features from adjacent patches). We took the logarithm of the depth map and performed learning and prediction in this space. The entire dataset contained more than 2 million data points ($107\times 86 \times 274$).  
We applied a Gaussian RBF kernel on the depth information, with the bandwidth parameter set to the median distance between training depths, and an approximation residual of $\epsilon=10^{-3}$. We used a linear kernel for the image features. 


\begin{figure}
  \centering
  \begin{tabular}{cc}
    \includegraphics[width=0.40\columnwidth]{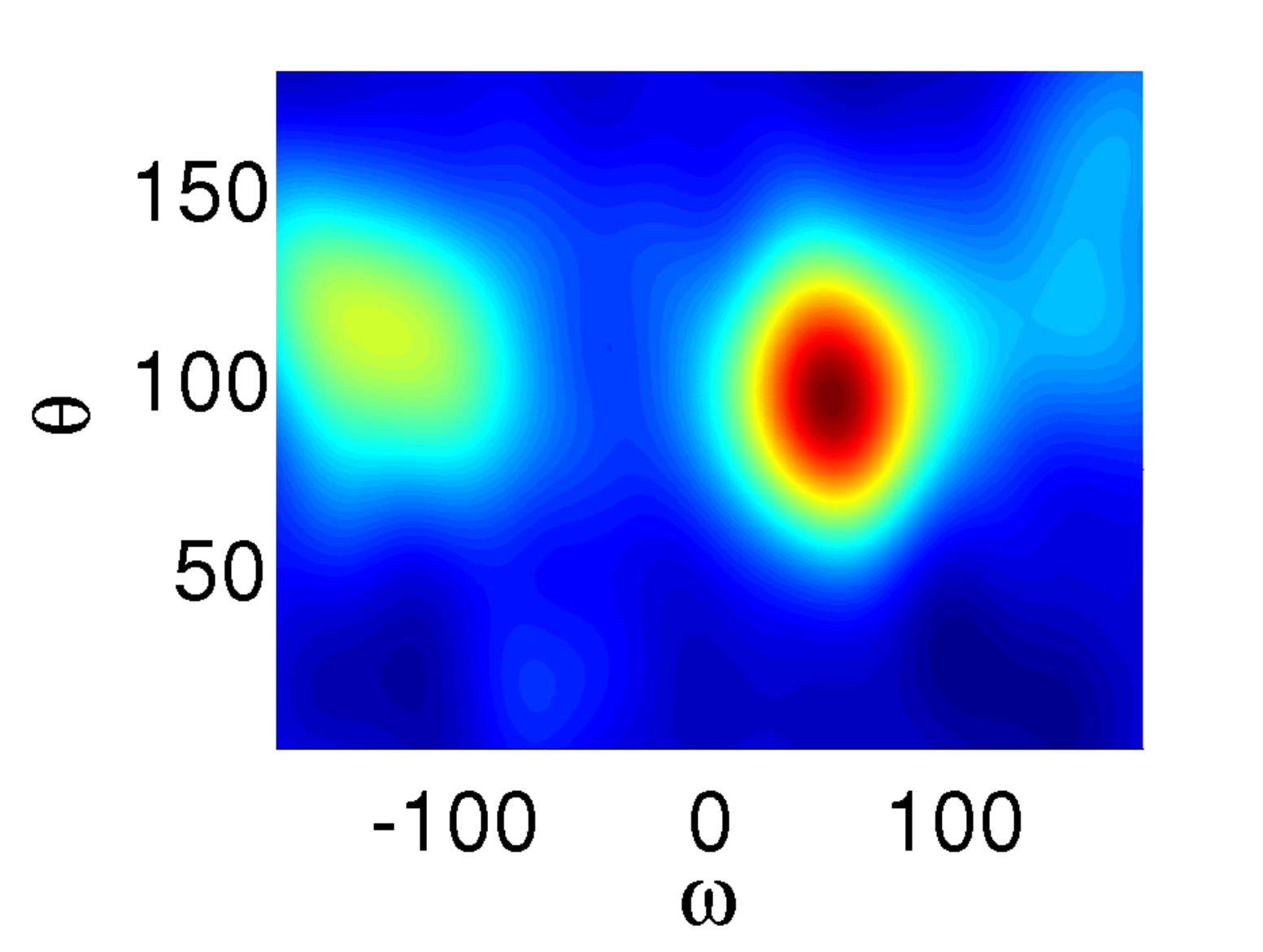} &
    \includegraphics[width=0.40\columnwidth]{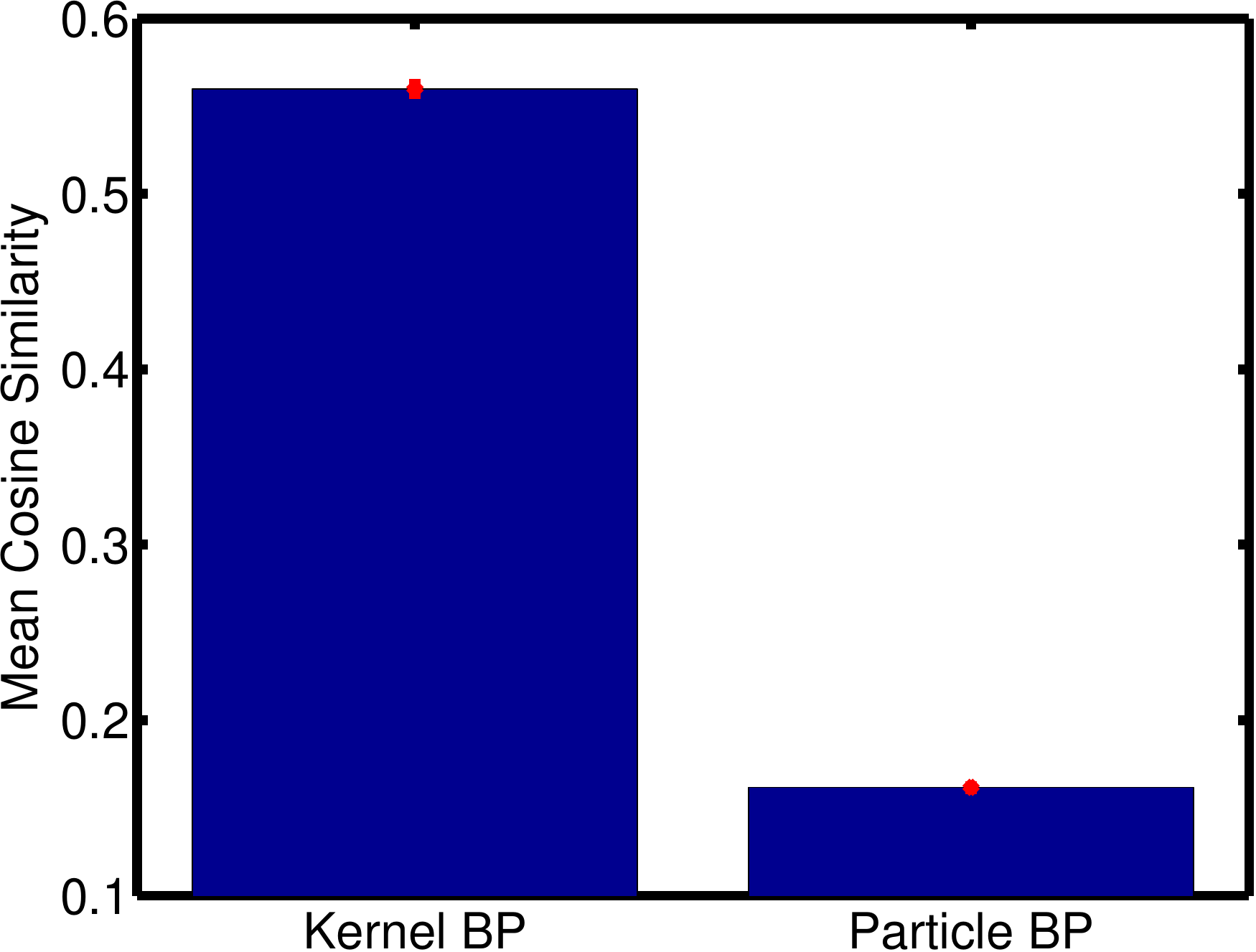} \\
    (a) {\footnotesize Angle distribution} & (b) {\footnotesize Prediction accuracy}
    \vspace{-3mm}
  \end{tabular}
  \caption{\footnotesize Average angle prediction accuracy of kernel versus particle BP in the protein folding problem.  
  }
  \label{fig:protein}
  \vspace{-5mm}
\end{figure} 

Our results were obtained by leave-one-out cross validation. 
For each test image, we ran discrete, Gaussian mixture, particle, and kernel BP for 10 BP iterations. The average prediction error (MAE: mean absolute error) and runtime are shown in Figures~\ref{fig:depth_results}(c) and (d). Kernel BP produces the lowest error (MAE=0.145) by a significant margin, while having a similar runtime to discrete BP. Gaussian mixture and particle BP achieve better MAE than discrete BP, but their runtimes are two order of magnitude slower. We note that the error of kernel BP is slightly better than the results of pointwise MRF reported in~\cite{SaxSunNg09}.





{\bf Protein structure prediction:} Our final experiment investigates the protein folding problem. 
The folded configuration of a  protein of length $n$ is roughly determined by 
a sequence of angle pairs $\{(\theta_i,\omega_i)\}_{i=1}^n$, each specific to an amino acid position. The goal is to predict the sequence of angle pairs given only the amino acid sequence. The two angles $(\theta_i,\omega_i)$ have ranges $[0,180]$ and $(-180,180]$ respectively, such that they correspond to points 
on the unit sphere $S^2$. Kernels yield an immediate solution to inference on these data:
\citet[Theorem 17.10]{Wendland05} provides a sufficient condition for a function on $S^2$ to be positive definite, satisfied by  $k(x,x'):=\exp(\sigma\inner{x}{x'})$, where $\inner{x}{x'}$ is the standard inner product between Euclidean coordinates.
%
Given the data are continuous, multimodal, and on a non-Euclidean domain (Figure~\ref{fig:protein}(a)), it is not obvious how 
Gaussian mixture or discrete BP might be applied. We therefore focus on comparing kernel and particle BP.

We obtained a collection of $1,400$ proteins with length larger than 100 from PDB. We first ran PSI-BLAST
to generate the sequence profile (a 20 dimensional feature for each amino acid position), and then used this profile as features for predicting the folding structure~\citep{Jones99}.
The graphical model was a chain of connected angle pairs, where each angle pair was associated with a 20 dimensional feature. 
We used a linear kernel on the sequence features. For the kernel between angles, the  bandwidth parameter was set at the median inner product between training points, and we used the  approximation residual $\epsilon=10^{-3}$. For particle BP, we learned the nonparametric potentials using $\exp(\sigma\inner{x}{x'})$ as the basis functions. 

In Figure~\ref{fig:protein}(b), we report the average prediction accuracy (Mean Cosine Similarity between the true coordinate $x$ and the predicted $x'$,~i.e.,~$\inner{x}{x'}$) over a 10-fold cross-validation process. In this case, kernel BP achieves a significantly better result than particle BP while running much faster (runtimes not shown due to space constraints).   





\vspace{-3mm}
\section{Conclusions and Further Work}
\vspace{-3mm}

We have introduced  {\em kernel belief propagation}, where the messages are functions in an RKHS. Kernel BP performs learning and inference on challenging graphical models with structured and continuous random variables, and is more accurate and much faster than earlier nonparametric BP algorithms. A possible extension to this work would be to kernelize tree-reweighted belief propagation~\citep{WaiJaaWill03b}. The convergence of kernel BP is a further challenging topic for future work~\citep{IhlFisWil05}.

{\small
{\bf Acknowledgements:} We thank Alex Ihler for the Gaussian mixture BP codes and helpful discussions. LS is supported by a Stephenie and Ray Lane Fellowship. This research was also supported by ARO MURI W911NF0710287, ARO MURI W911NF0810242,
NSF Mundo IIS-0803333, NSF Nets-NBD CNS-0721591
and ONR MURI N000140710747.}


\clearpage
\newpage
\onecolumn
\section*{Supplementary to Kernel Belief Propagation}
\setcounter{section}{0} 
Section \ref{sec:GMandPBP_supp} contains a review of Gaussian mixture BP and particle BP, as well as a detailed explanation of our strategy for learning edge potentials for these approaches from training data.
Section \ref{sec:discreteAndParticle_supp} provides  parameter settings and experiment details for particle BP and discrete BP, in the synthetic image denoising and depth reconstruction experiments.
Section \ref{sec:approxMessageUpdates_supp} contains a comparison of two different approximate feature sets: low rank approximation of the tensor features and low rank approximation of the individual features alone.
Section \ref{sec:paperCategory_supp} is an experiment on learning paper categories using citation networks.
Sections \ref{sec:localMarginalConsistency_supp} and \ref{sec:BPusingLearnedPotentials_supp} demonstrate the optimization objective of locally consistent BP updates, and 
provide a derivation of these updates in terms of the conditional expectation. 
Section \ref{sec:kernelGaussianBP_supp}
discusses the kernelization of Gaussian BP.
Section \ref{sec:messageError_supp} gives the error introduced by low rank approximation of the 
messages.

\section{Gaussian Mixture and Particle BP}\label{sec:GMandPBP_supp}

We describe two competing approaches for nonparametric
belief propagation: Gaussian mixture BP, originally known as non-parametric BP \citep{SudIhlFreWil03}, and
particle BP \citep{IhlMcA09}. For these algorithms, the edge potentials $\Psi(x_{s},x_{t})$, self-potentials $\psi(x_{t})$,
and evidence potentials $\Psi(x_{t},y_{t})$ must be provided in advance
by the user. Thus, we begin by describing how the edge potentials
in Section 2 of the main document 
may be learned from training data, but in a form applicable
to these inference algorithms: we express $\mathbb{P}(x_t|x_s)$ as a mixture of Gaussians.
We then describe the inference algorithms themselves.

In learning the edge potentials, we turn to \cite{SugTakSuzKanetal10},
who provide a least-squares estimate of a conditional density in the
form of a mixture of Gaussians,\[
\mathbb{P}(v|u)=\sum_{i=1}^{b}\alpha_{i}\kappa_{i}(u,v)=\alpha^{\top}\pmb{\kappa}_{u,v},\]
where $\kappa_{i}(u,v)$ is a Gaussian with diagonal covariance centred
at%
\footnote{These centres may be selected at random from the training observations.
We denote the mixture kernel by $\kappa(u,v)$ to distinguish it from the RKHS
kernels used earlier.%
} $(q_{i},r_{i})$. Given a training set $\{(u^{j},v^{j})\}_{j=1}^{m}$,
we obtain the coefficients \[
\alpha:=\left[\left(\widehat{H}+\lambda I\right)^{-1}\hat{h}\right]_{+},\]
where $\widehat{H}:=\sum_{j=1}^{m}\int_{\mathcal{V}}\pmb{\kappa}_{u^{j},v}\pmb{\kappa}_{u^{j},v}^{\top}dv$,
$\hat{h}:=\sum_{j=1}^{m}\pmb{\kappa}_{u^{j},v^{j}}$, $\lambda$ is
a regularization coefficient, and $[\alpha]_{+}$ sets all negative
entries of its argument to zero (the integral in $\widehat{H}$ can
easily be computed in closed form). We emphasize that the  Gaussian mixture
 representation takes a quite different form to the RKHS
representation of the edge potentials. 
Finally, to introduce
evidence, we propose to use kernel ridge regression to provide a mean
value of the hidden variable $x_{t}$ given the observation $y_{t}$,
and to center a Gaussian at this value: again, the regression function
is learned nonparametrically from training data.

We now describe how these edge potentials are incorporated into Gaussian
mixture BP. Assuming the incoming messages are each a mixture of $b$
Gaussians, the product of $d_{t}$ such messages will contain $b^{d_{t}}$
Gaussians, which causes an exponential blow-up in representation size
and computational cost. In their original work, Sudderth et al. address this issue using an
approximation scheme. First, they subsample from the incoming mixture
of $b^{d_{t}}$ Gaussians to draw $b$ Gaussians, at a computational
cost of $O(d_{t}\tau b^{2})$ \emph{for each node}, where $\tau$
is the number of iterations of the associated Gibbs sampler (see their
Algorithm 1). The evidence introduced via kernel ridge regression
is then incorporated, using a reweighting described by their Algorithm
2. Finally, in Algorithm 3, $b$ samples $\left\{ x_{t}^{i}\right\} _{i=1}^{b}$
are drawn from the reweighted mixture of $b$ Gaussians, and for each
of these, $\left\{ x_{s}^{i}\right\} _{i=1}^{b}$ are drawn from
the conditional distribution $x_{s}|x_{t}^{i}$ arising from the
edge potential $\psi(x_{s},x_{t})$ (which is itself a Gaussian mixture,
learned via the approach of Sugiyama et al.). Gaussians are placed
on each of the centres $\left\{ x_{s}^{i}\right\} _{i=1}^{b}$,
and the process is iterated.

In our implementation, we used the more efficient multiscale KD-tree sampling method of \cite{KD-tree}. We converted the Matlab Mex implementation of \cite{KDEToolbox}
to C++, and used GraphLab to execute sampling in parallel with up to 16 cores. An input parameter to the sampling procedure is $\epsilon$, the level of accuracy. We performed a line search to set $\epsilon$ for high accuracy, but limited the execution time to be at most 1000 times slower than KBP. 

Finally, we describe the inference procedure performed by Particle
BP. In this case, each node $t$ is associated with a set of particles
$\left\{ x_{t}^{i}\right\} _{i=1}^{b}$, drawn i.i.d. from a distribution
$W_{t}(x_{t})$. Incoming messages $m_{ut}$ are expressed as weights
of the particles $x_{t}^{i}$. Unlike Gaussian mixture BP, the incoming
messages all share the same set of particles, which removes the need
for Parzen window smoothing. The outgoing message $m_{ts}$ is computed
by summing over the product of these weights and the edge and evidence
potentials at the particles, yielding a set of weights over samples
$\left\{ x_{s}^{i}\right\} _{i=1}^{b}$ at node $s$; the procedure
is then iterated \citep[see][eq. 8]{IhlMcA09}. 
We again implement
this algorithm using edge potentials computed according to Sugiyama
et al. Since an appropriate sample distribution $W_{t}$ is hard to
specify in advance, a resampling procedure must be carried out at
each BP iteration, to refresh the set of samples at each node and
ensure the samples cover an appropriate support (this is a common
requirement in particle filtering). Thus, each iteration of Particle
BP requires a Metropolis-Hastings chain to be run for every node,
which incurs a substantial computational cost.
That said, we found that in practice, the resampling could be conducted less often
 without an adverse impact on performance, but resulting in major improvements in runtime,
 as described in Section \ref{sec:discreteAndParticle_supp} below.
 See \cite[Section 6]{IhlMcA09}
for more detail.



\section{Settings for Discrete and Particle BP}\label{sec:discreteAndParticle_supp}

\subsection{Depth Reconstruction from 2-D Images}
\subsubsection{Discrete BP}
The log-depth was discretized into 30 bins, and edge parameters were
 selected
to achieve locally consistent Loopy BP marginals using the technique described
in \cite{WaiJaaWill03b}. Empirically, finer discretizations did not improve
resultant accuracy, but increased runtime significantly. We used the
Splash scheduling of \cite{GonLowGue09} since it provided the lowest runtime
among all tested schedulings.

\subsubsection{Particle BP}
The particle belief propagation implementation was particularly difficult to
tune due to its excessively high runtime. Theoretically, results comparable to
the Kernel BP method were attainable. However in practice, the extremely high
cost of the resampling phase on large models meant that only a small number of
particles could be maintained if a reasonable runtime was to be achieved on our
evaluation set of 274 images.

Ultimately, we decided to find a configuration which allowed us to complete
the evaluation in about 2 machine-days on an 8-core Intel Nehalem machine;
allowing inference on each evaluation image to take 10 minutes of parallel
computation. For each image, we ran 100 iterations of a
simple linear-sweep scheduling, using 20 particles per message, and
resampling every 10 iterations. Each resampling phase ran MCMC for a maximum of 10 steps per particle.
We also implemented acceleration tricks
where low weight particles ($<1E-7$ after normalization) were ignored during the message
passing process. Empirically this decreased runtime without affecting the quality of results.

\subsection{Synthetic Image Denoising}
\subsubsection{Discrete BP}
To simplify evaluation, we permitted a certain degree of ``oracle'' information,
by matching the discretization levels during inference with the color levels
in the ground-truth image.

We evaluated  combined gradient/IPF + BP methods here to
learn the edge parameters. We found that gradient/IPF performed well when there
were few colors in the image, but failed to converge when the number of
colors increased into the hundreds. This is partly due to the instability of BP,
as well as the large number of free parameters in the edge potential.

Therefore once again, edge potentials were selected using the technique
described in \cite{WaiJaaWill03b}. This performed quite well empirically, as
seen in Figure 1(c) (main document).

\subsubsection{Particle BP}
The high runtime of the Particle Belief Propagation again made accuracy
evaluation difficult. As before, we tuned the particle BP parameters to
complete inference on the evaluation set of 110 images in 2 machine days,
allowing about 25 minutes per evaluation image. We ran 100 iterations of
30 particles per message, resampling every 15 iterations.
Each resampling phase ran MCMC for a maximum of 10 steps per particle.


\section{Effects of Approximate Message Updates}\label{sec:approxMessageUpdates_supp}




In this section, we study how different levels of feature approximation error $\epsilon$ affect the speed of kernel BP and the resulting performance.
Our experimental setup was the image denoising experiment described in Section 5.1 of the main document.
 We note that the computational cost of our constant message update is $O(\ell^2 d_{\max})$ where $\ell$ is inversely related to the approximation error $\epsilon$. This is a substantial runtime improvement over naively applying a low rank kernel matrix approximation, which only results in a linear time update with computational cost $O(\ell m d_{\max})$. In this experiment, we varied the feature approximation error $\epsilon$ over three levels,~\ie~$10^{-1},10^{-2},10^{-3}$, and compared both speed and denoising performance of the constant time update to the linear time update. 
 
\begin{figure}[t]
  \hspace{-1cm}
  \centering
  \begin{tabular}{cc}
    \includegraphics[width=0.45\columnwidth]{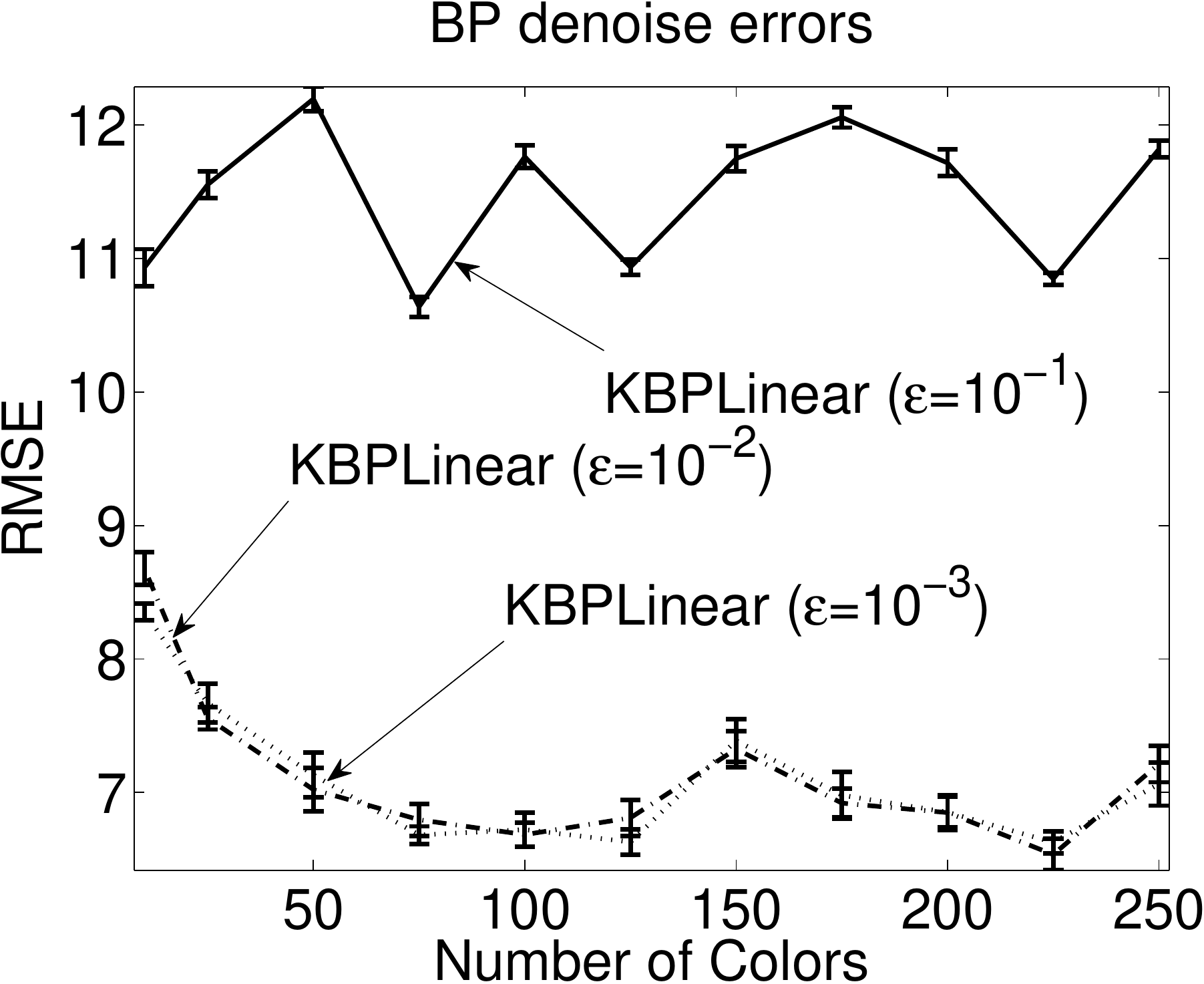} &
    \includegraphics[width=0.45\columnwidth]{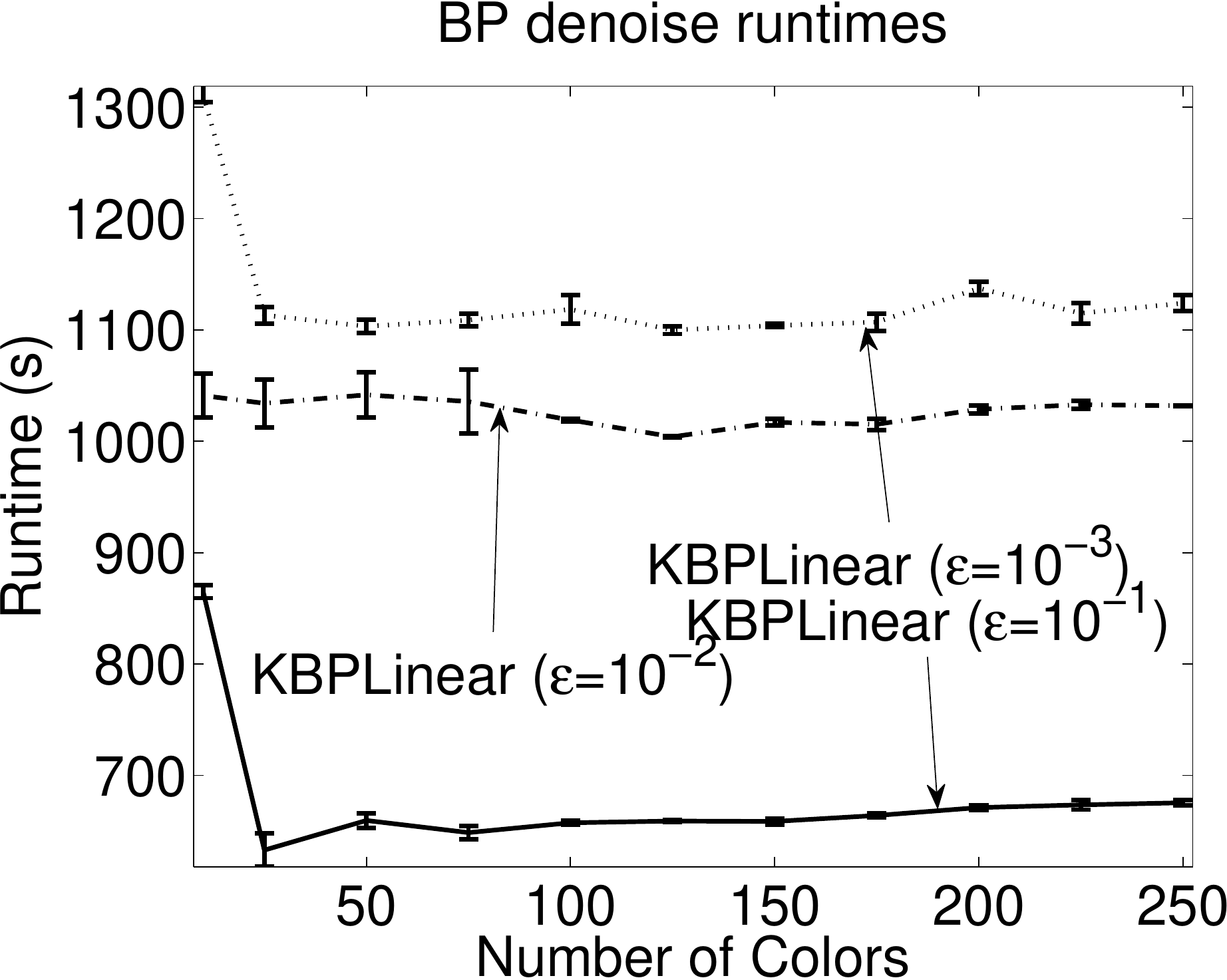}  \\
    (a) Error (linear time approximation) & (b) Runtime (linear time approximation) \\    
    \includegraphics[width=0.45\columnwidth]{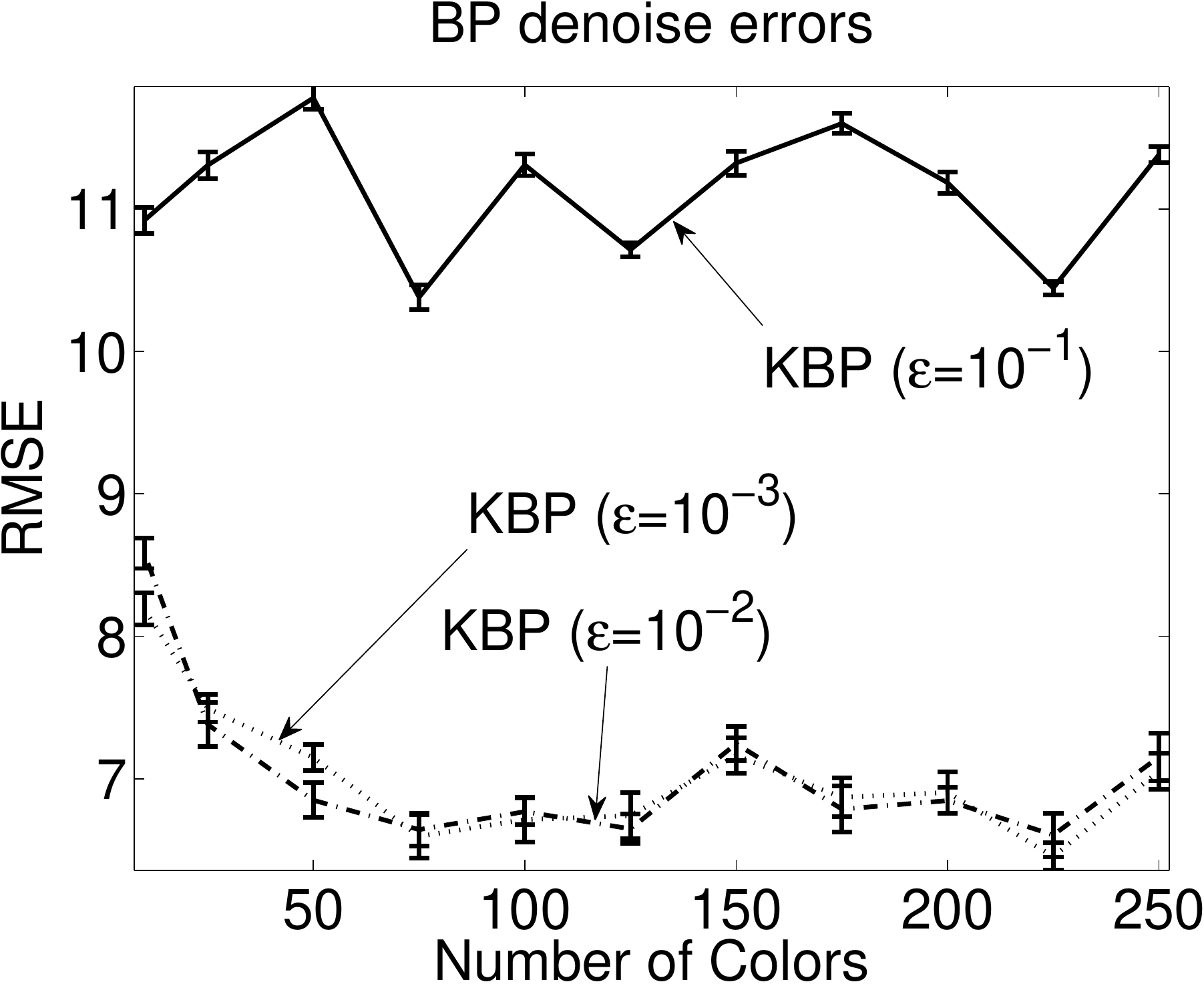} &
    \includegraphics[width=0.45\columnwidth]{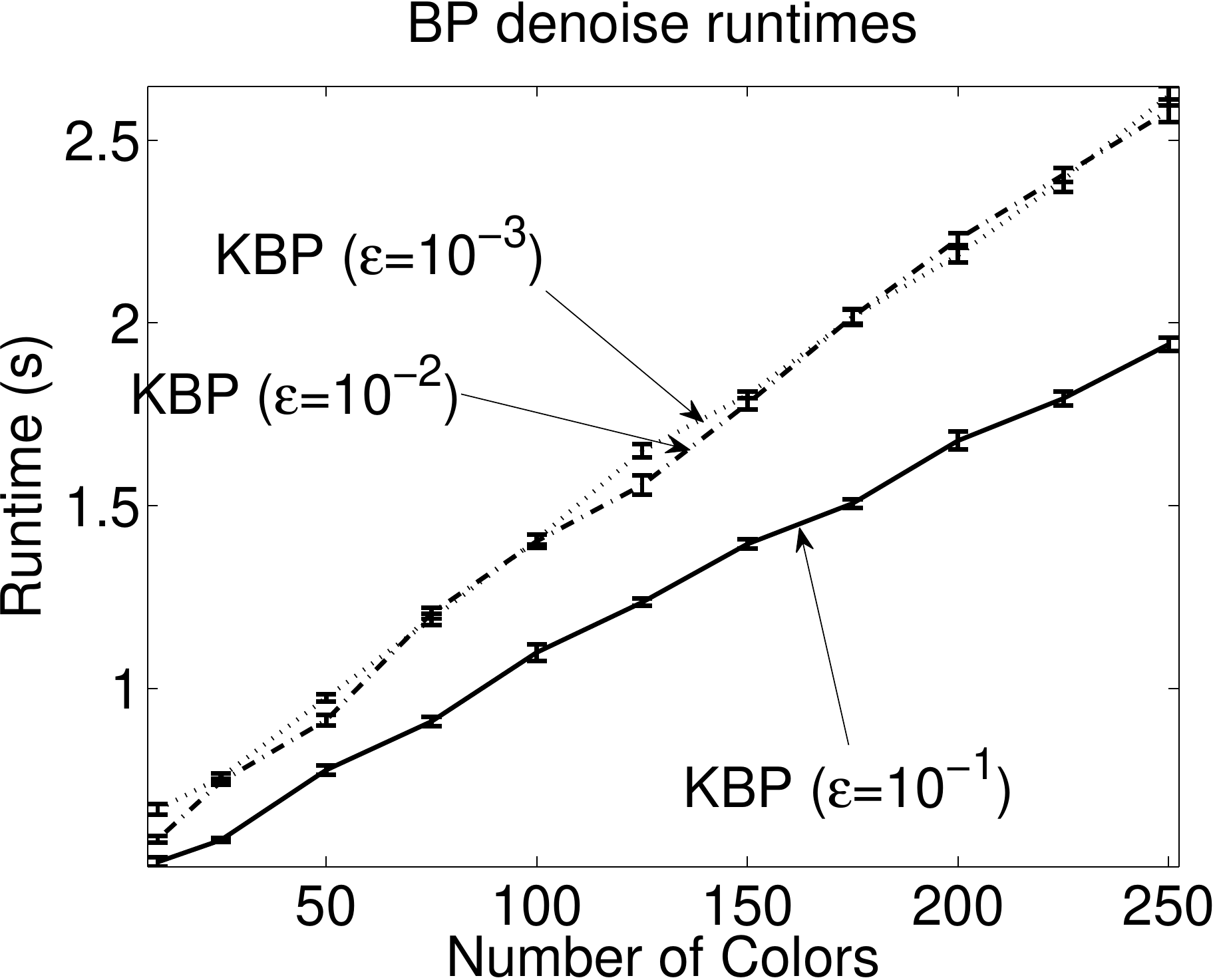} 
     \\
    (c) Error (constant time approximation) & (d) Runtime (constant time approximation) \\        
  \end{tabular}
  \caption{Average denoising error and runtime of linear time kernel BP versus constant time kernel BP, using different feature approximation errors, over 10 test images with a varying number of image colors.}
  \label{fig:approx_level}
\end{figure}

From Figures~\ref{fig:approx_level} (a) and (c), we can see that for each approximation level, the constant time update achieves about the same denoising performance as the linear time update, while at the same time being orders of magnitude faster (by comparing Figures~\ref{fig:approx_level} (b) and (d)). Despite the fact that the constant time update algorithm makes an additional approximation to the tensor product features, its denoising performance is not affected. We hypothesize that the degradation in performance is largely caused by representing the messages in terms of a small number of basis functions, while the approximation to the tensor features introduces little additional degradation. 

\begin{figure}[t]
  \hspace{-1cm}
  \centering
  \begin{tabular}{cc}
    \includegraphics[width=0.45\columnwidth]{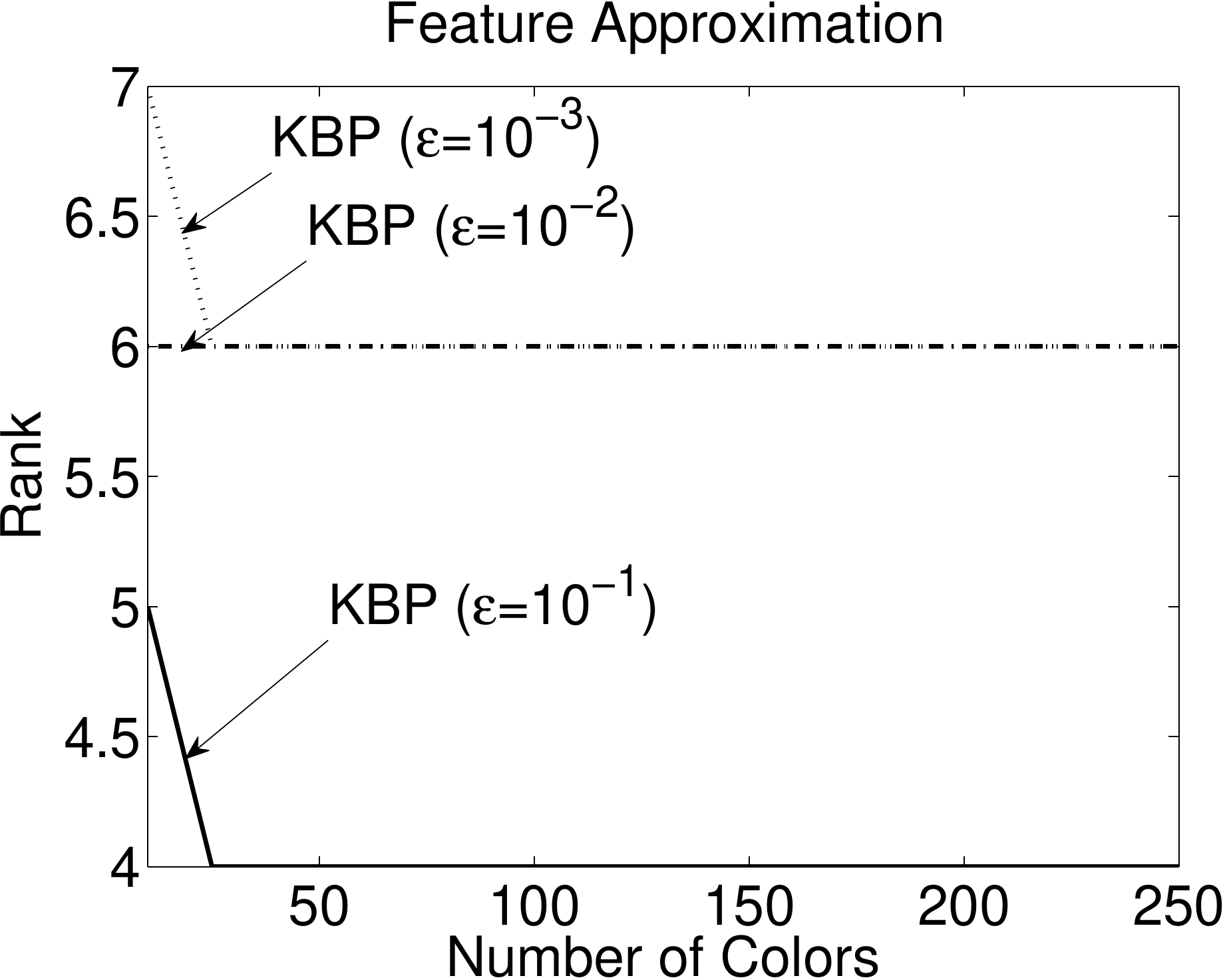} &
    \includegraphics[width=0.45\columnwidth]{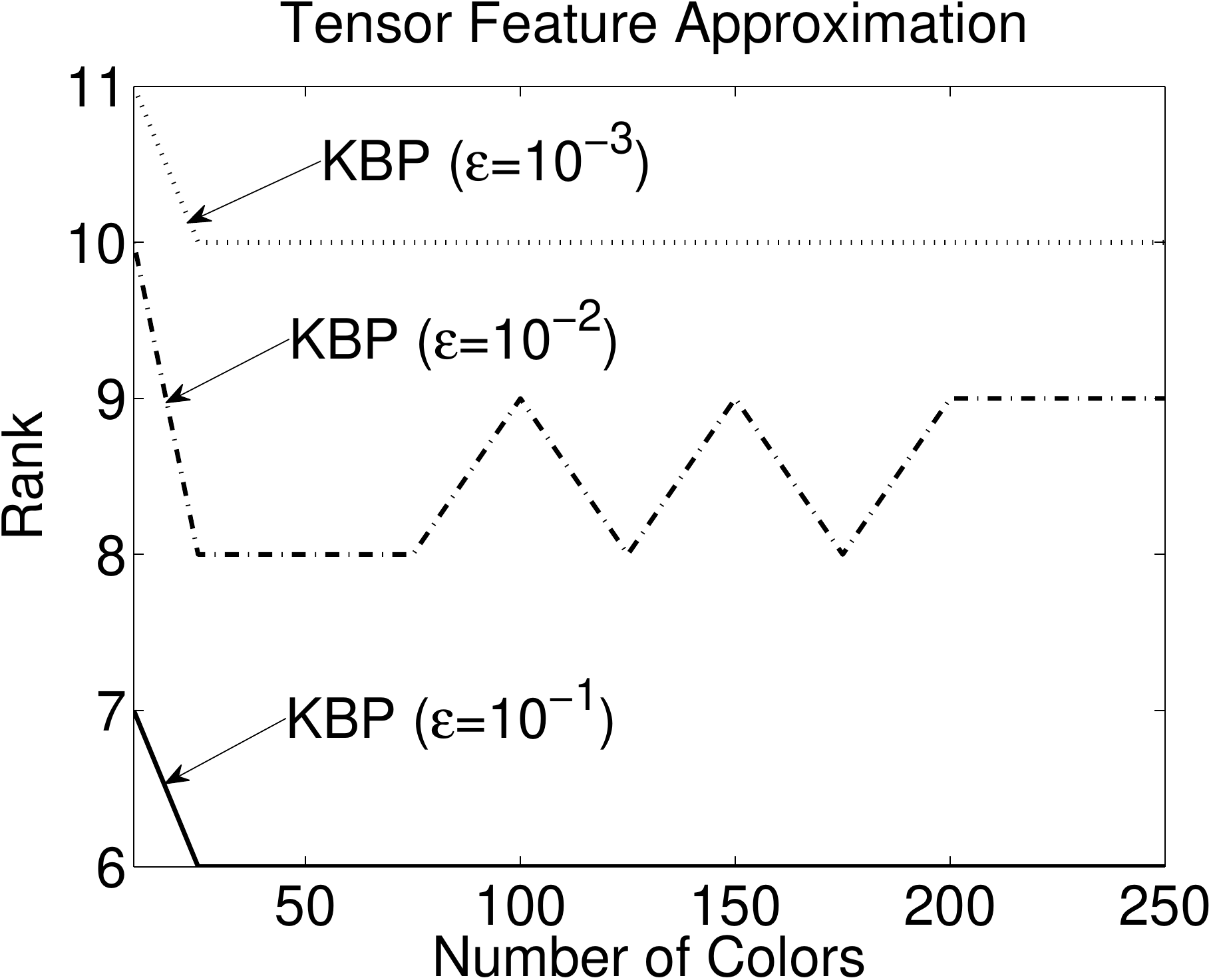}  \\
    (a) Rank for feature approximation & (b) Rank for tensor feature approximation \\    
  \end{tabular}
  \caption{The rank obtained for kernel feature approximation and tensor feature approximation for different levels of approximation error $\epsilon$, for the image denoising problem.}
  \label{fig:approx_rank}
\end{figure}

Another interesting observation from Figure~\ref{fig:approx_level} (d) is that the runtime of constant time kernel BP update increases as the number of colors in the image increases. This is mainly due to the increased number of test points as the color number increases; and also partially due to the increased  rank needed for approximating the tensor features. In Figure~\ref{fig:approx_rank}, we plot the rank needed for kernel feature approximation and tensor feature approximation for different numbers of colors and different approximation errors $\epsilon$. It can be seen that in general, as we use a smaller approximation error, the rank increases, leading to a slight increase in runtime.   

Finally, we compare with kernel belief propagation in the absence of any low rank approximation (KBP Full). Since KBP Full is computationally expensive, we reduced the denoising problem to images of size $50\times 50$ to allow KBP to finish in reasonable time. We only compared on  100 color images, again for reasons of cost. We varied the feature approximation error for the constant time and linear time approximation over three levels, $10^{-1}$, $10^{-2}$, $10^{-3}$, and compared
both speed and denoising performance of KBP Full versus the constant time and linear time updates.  

\begin{figure}[h]
  \hspace{-1cm}
  \centering
  \begin{tabular}{cc}
    \includegraphics[width=0.45\columnwidth]{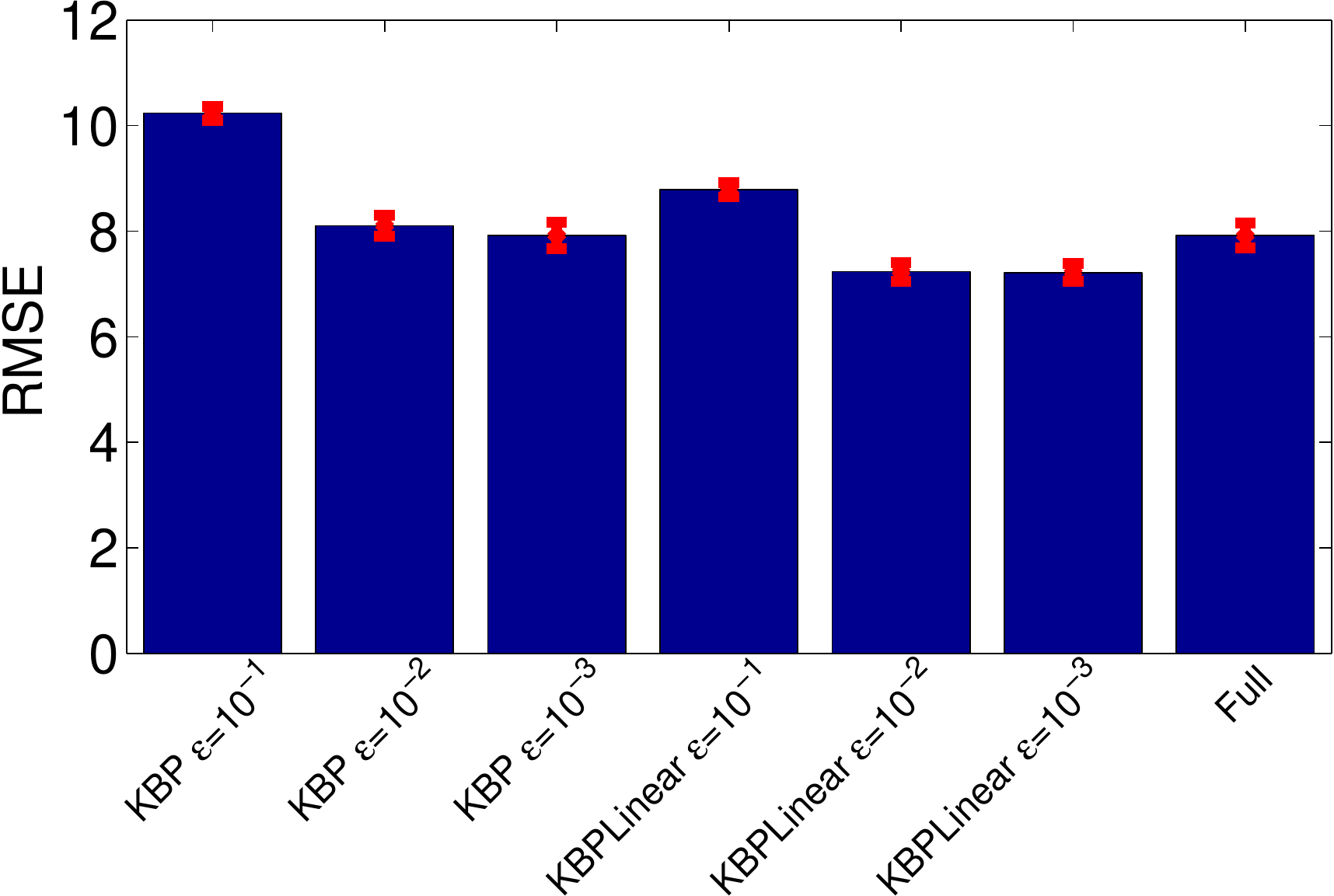} &
    \includegraphics[width=0.47\columnwidth]{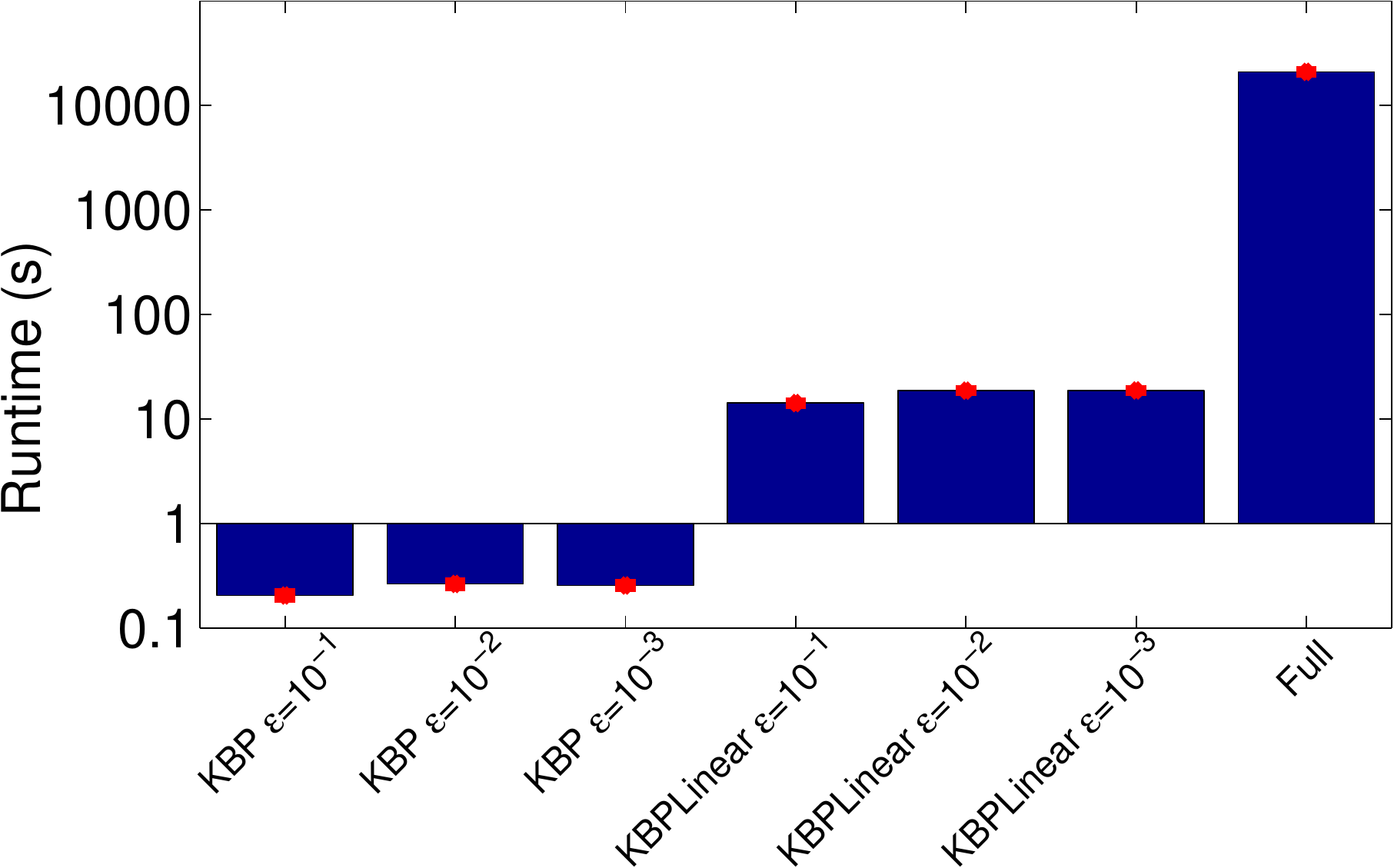}  \\
    (a) Error & (b) Runtime (vertical axis is in log scale) \\    
  \end{tabular}
  \caption{(a) Denoising error for the constant time approximate update (KBP) and linear time approximate update (KBPLinear), over three levels of approximation error $\epsilon=\{10^{-1},10^{-2},10^{-3}\}$, versus  kernel BP without low rank approximation (Full). (b) Runtime corresponding to different approximation schemes.}
  \label{fig:denoise50}
\end{figure}

The comparisons are shown in Figure~\ref{fig:denoise50}. We can see from Figure~\ref{fig:denoise50}(a) that the denoising errors for constant time and linear time approximations decrease as we decrease the approximation error $\epsilon$. Although the denoising error of KBP Full is slightly lower than constant time approximations, it is a slight increase over the linear time approximation at $\epsilon=10^{-3}$. One reason might be that the kernel approximation also serves as a means of regularization when learning the conditional embedding operator. This additional regularization may have  slightly improved the generalization ability of the linear time approximation scheme. In terms of runtime (Figure~\ref{fig:denoise50}(b)), constant time approximation is substantially faster than linear time approximation and KBP Full. In particular, it is nearly 100 times faster than the linear time algorithm, and 10000 times faster than KBP Full. 

\section{Predicting Paper Categories}\label{sec:paperCategory_supp}



In this experiment, we predict paper categories from a combination of the paper features and their citation network.
Our data were obtained by crawling 143,086 paper abstracts from the ACM digital library, and  extracting the citation networks linking these papers. Each paper was labeled with a variable number of categories, ranging from 1 to 10; there were a  total of 367 distinct categories in our dataset. For simplicity, we ignored directions in the citation network, and treated it as an undirected graph (i.e, we did not distinguish ``citing'' and ``being cited''). The citation network was sparse, with more than 85\% of the papers having fewer than 10 links. The maximum number of links was 450. 

Paper category prediction is a multi-label problem with a large output space. The output is very sparse: the label vectors have only a small number of nonzero entries. In this case, the simple one-against-all approach of learning a single predictor for each category can become prohibitively expensive, both in  training and in testing. Recently, \cite{HsuKakLanZha09} proposed to solve this problem using compressed sensing techniques: high dimensional sparse category labels are first compressed to lower dimensional real vectors using a random projection, and  regressors are learned for these real vectors. In the testing stage, high dimensional category labels are decoded from the predicted real vectors of the test data using orthogonal marching pursuit (OMP). 

For the purposed of the present task, however, the compressed sensing approach ignores information from the citation network: papers that share categories tend to cite each other more often. Intuitively, taking into account category information from neighboring papers in the citation network should improve the performance of category prediction. This intuition can be formalized using undirected graphical models: each paper $i$ contains a category variable $y_i\in\{0,1\}^{367}$, and these variables are connected according to the citation network; each category variable is also connected to a variable $x_i$ corresponding to the abstract of the paper. In our experiment, we used 9700 stem words for the abstracts, and $x_i\in\RR^{9700}$ was the tf-idf vector for paper $i$. The graphical model thus contains two types of edge potential, $\Psi(y_i,y_i)$ and $\Psi(y_i, y_k)$, where $k\in\Ncal(j)$ is the neighbor of $j$ according to the citation network. 

It is difficult to learn this  graphical model and perform inference on it, since the category variables $y_i$ have high cardinality, making the marginalization step in BP prohibitively expensive. Inspired by the compressed sensing approach for multilabel prediction, we first employed random projection kernels, and then used our kernel BP algorithm. 
Let $A\in\RR^{d\times367}$ be a random matrix containing~\iid~Gaussian random variables of zero mean and variance $1/d$. We defined the random projection kernel for the category labels to be $k(y,y') = \inner{Ay}{Ay'} = \inner{\phi(y)}{\phi(y')}$, and used a linear kernel for the abstract variables. We ran kernel BP for 5 iterations, since further iterations did not improve the performance. MAP assignment based on the belief was performed by finding a unit vector $\phi(\hat{y})= A\hat{y}$ that maximized the belief. The sparse category labels $\hat{y}$ were decoded from $\phi(\hat{y})$ using OMP. 
\begin{figure}
    \centering
    \includegraphics[width=0.50\columnwidth]{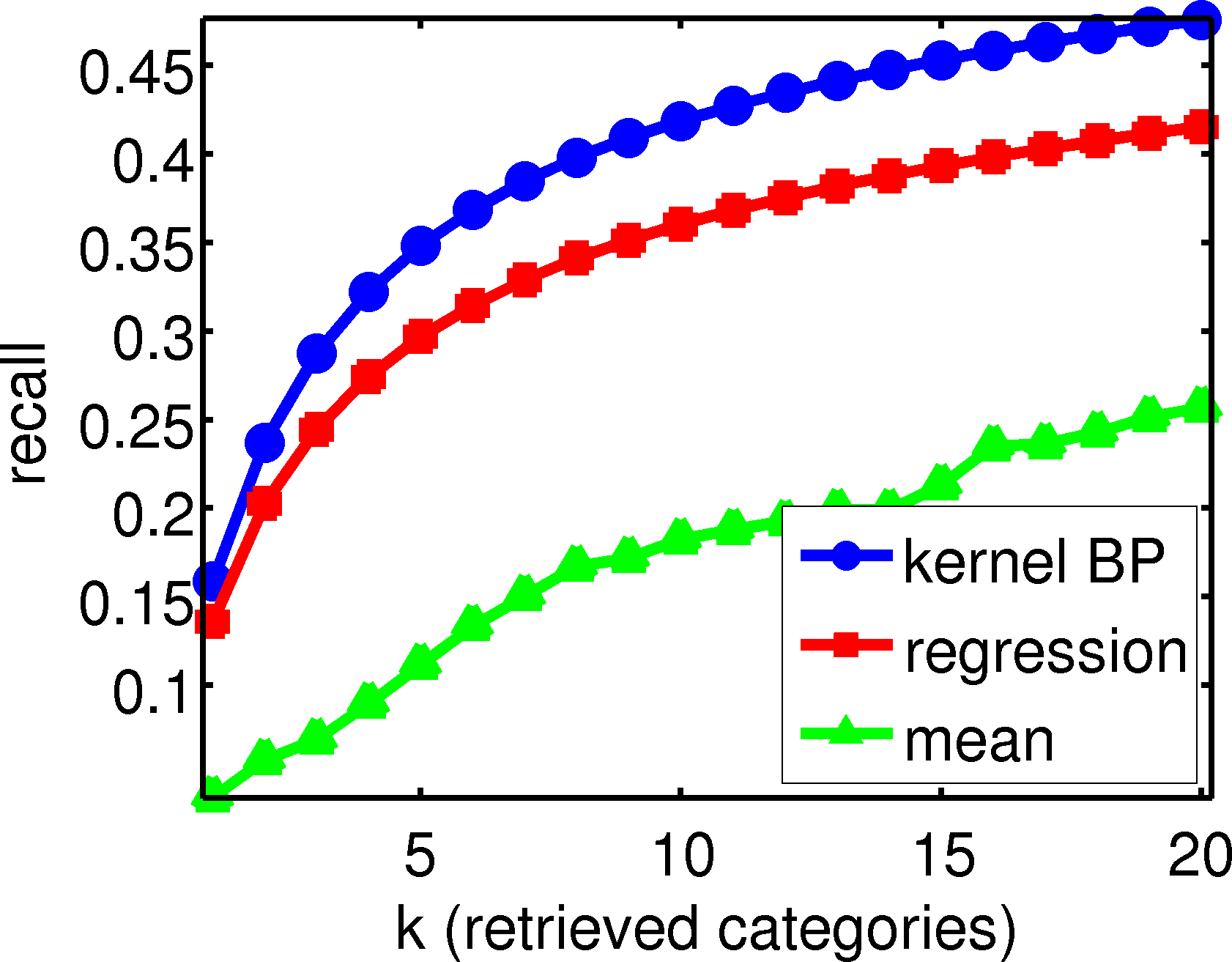}
    \caption{Comparison of kernel BP, multilabel prediction via compressed sensing (regression), and a baseline prediction using the top $k$ most frequent categories (mean) for ACM paper category prediction. 
  }   
  \label{fig:acm_results}    
\end{figure}

To measure experimental performance, we performed 10 random splits of the papers, where in each split we used 1/3 of the papers for training and the remaining 2/3 for testing. The random splits were controlled in such a way that high degree nodes (with degree $> 10$) in the citation networks always appeared in the training set. Such splitting  reflects the data properties expected in real-world scenarios:   important papers (high degree nodes which indicate either influential papers or survey papers) are usually  labeled,
whereas the majority of  papers may not have a label; the automatic labeling is mainly
 needed for these less prominent papers.
We used recall@$k$ in evaluating the performance of our method on the test data. 
We compared against the regression technique of Hsu et al. for multilabel prediction, and a
 baseline prediction using the top $k$ most frequent categories. 
For both our method and the method of Hsu et al., we used a random projection matrix with $d=100$. 

Results are shown in Figure~\ref{fig:acm_results}. Kernel BP  performs better than multilabel prediction via compressed sensing (i.e., the independent regression approach, which ignores graphical model structure) over a range of $k$ values. In particular, for the top 10 and 20 predicted categories, kernel BP achieves  recall scores of 0.419 and 0.476, respectively, as opposed to 0.362 and 0.417 for  independent regression. 

\section{Local Marginal Consistency Condition When Learning With BP}\label{sec:localMarginalConsistency_supp}

In this section, we show that fixed points of BP satisfy particular marginal consistency conditions (\ref{eq:relation_marginal0}) and (\ref{eq:relation_marginal}) below. As we will see, these arise from the fact that we are using a Bethe free energy approximation in fitting our model, and the form of the fixed point equations that define the minimum of the Bethe free energy. The material in this section draws from a number of references \citep[for instance][]{YedFreWei01,YedFreWei05,WaiJor08,KolFri09}, but is presented in a form specific to our case, since we are neither in a discrete domain nor using exponential families.

The parameters of a pairwise Markov random field (MRF) can be learned by maximizing the log-likelihood of the model $\PP$ with respect to true underlying distribution $\PP^\star$. Denote the model by 
$$\PP(\bm{X}):=\frac{1}{Z} \prod_{(s,t)\in\mathcal{E}}\Psi_{st}(X_{s},X_{t})\prod_{s\in\mathcal{V}}\Psi_{s}(X_{s})$$
 where $Z:=\int_{\bm{X}} \prod_{(s,t)\in\mathcal{E}}\Psi_{st}(X_{s},X_{t})\prod_{s\in\mathcal{V}}\Psi_{s}(X_{s})$ is the partition function that normalizes  the distribution. The model parameters $\cbr{\Psi_{st}(X_{s},X_{t}), \Psi_{s}(X_{s})}$ can be estimated by maximizing 
\begin{align}
  \Lcal 
  &= \EE_{\bm{X}\sim \PP^\star(\bm{X})} \sbr{\log \PP(\bm{X})} \nonumber \\
  &= \EE_{\bm{X}\sim \PP^\star(\bm{X})} \sbr{ \sum_{(s,t)\in\Ecal} \log \Psi_{st}(X_s,X_t) + \sum_{s \in \Vcal} \log \Psi_{s}(X_s) - \log Z}.  
\end{align}
Define $\widetilde{\Psi}_{st}(X_{s},X_{t}):=\log\Psi_{st}(X_{s},X_{t})$ and $\widetilde{\Psi}_{s}(X_{s}):=\log \Psi_{s}(X_{s})$. Setting the derivatives of $\Lcal$ with respect to $\cbr{\widetilde{\Psi}_{st}(X_{s},X_{t}), \widetilde{\Psi}_{s}(X_{s})}$ to zero, we have
\begin{align}
  \frac{\partial \Lcal}{\partial \widetilde{\Psi}_{s}(X_{s})} 
  =&~\PP^\star(X_s,X_t) - \frac{\partial \log Z}{{\partial \widetilde{\Psi}_{st}(X_{s},X_{t})}}  = 0, \\
  \frac{\partial \Lcal}{\partial \widetilde{\Psi}_{s}(X_{s})} 
  =&~\PP^\star(X_s) - \frac{\partial \log Z}{{\partial \widetilde{\Psi}_{s}(X_{s})}} = 0.
\end{align}
For a general pairwise MRF on a loopy graph, computing the log-partition function, $\log Z$, is intractable. Following e.g. \cite{YedFreWei01,YedFreWei05} and \citet[Ch. 11]{KolFri09}, $\log Z$  may be approximated as a minimum of the  Bethe free energy with respect to a new set of parameters $\cbr{b_{st},b_s}$,    
\begin{align}
  F(\cbr{b_{st},b_s})
  =& \sum_{(s,t)\in\Ecal} \int_{\Xcal}\int_{\Xcal} b_{st}(X_s,X_t)\sbr{\log b_{st}(X_s,X_t) - \log \Psi_{st}(X_{s},X_{t}) \Psi_{s}(X_{s}) \Psi_{t}(X_{t})} dX_s dX_t \nonumber \\
   & - \sum_{s\in\Vcal} (d_s-1) \int_{\Xcal} b_s(X_s) \sbr{\log b_s(X_s) - \log \Psi_{s}(X_s)} dX_s,
\end{align}
subject to normalization and marginalization constraints, $\int_{\Xcal} b_s(X_s) dX_s = 1$, $\int_{\Xcal} b_{st}(X_s, X_t) dX_s = b_t(X_t)$. Let $F^\star:= \min_{\cbr{b_{st},b_s}} F$ (we note that Bethe free energy is not convex, and hence there can be multiple local minima. Our reasoning does not require constructing a specific local minimum, and therefore we simply write $F^\star$). The zero gradient conditions on the partial derivatives of $\Lcal$ are then approximated as
\begin{align}  
  \frac{\partial \Lcal}{\partial \widetilde{\Psi}_{s}(X_{s})} 
  \approx &~\PP^\star(X_s,X_t) - \frac{\partial F^\star}{{\partial \widetilde{\Psi}_{st}(X_{s},X_{t})}} = 0, \label{eq:gradientcond1} \\
  \frac{\partial \Lcal}{\partial \widetilde{\Psi}_{s}(X_{s})} 
  \approx &~\PP^\star(X_s) - \frac{\partial  F^\star}{\partial \widetilde{\Psi}_{s}(X_{s})} = 0.\label{eq:gradientcond2} 
\end{align}
Since $F$ is a linear function of $\cbr{\widetilde{\Psi}_{st}(X_{s},X_{t}), \widetilde{\Psi}_{s}(X_{s})}$ for every fixed $\cbr{b_{st},b_s}$, Danskin's
theorem  \cite[p. 717]{Bertsekas99} gives us a way to compute the partial derivatives of $F^\star$. These are 
\begin{align}
  \frac{\partial F^\star}{\partial \widetilde{\Psi}_{st}(X_{s},X_{t})} 
  =&~\frac{\partial F(\cbr{b_{st}^\star,b_s^\star})}{\partial \widetilde{\Psi}_{st}(X_{s},X_{t})} = b_{st}^\star(X_s,X_t), \\
  \frac{\partial  F^\star}{{\partial \widetilde{\Psi}_{s}(X_{s})}}
  =&~\frac{\partial F(\cbr{b_{st}^\star,b_s^\star})}{\partial \widetilde{\Psi}_{s}(X_{s})} = b_s^\star(X_s), 
\end{align}
where $\cbr{b_{st}^\star,b_s^\star}:=\argmin_{\cbr{b_{st},b_s}} F$. Therefore, according to~\eq{eq:gradientcond1}~and~\eq{eq:gradientcond2}, learning a pairwise MRF using the Bethe energy variational approximation to the log partition function results in the following matching conditions,
\begin{align}
  \PP^\star(X_s,X_t) =&~ b_{st}^\star(X_s,X_t), \\
  \PP^\star(X_s) =&~b_s^\star(X_s).  
\end{align}
We now introduce the notion of belief propagation as a means of finding the minima of the Bethe free energy. 
This will in turn lead to local marginal consistency conditions for learning with BP.
 \cite{YedFreWei01} showed that the fixed point of $F$  (and therefore the global minimum $\cbr{b_{st}^\star,b_s^\star}$)  must satisfy the relations
\begin{align}
  b_{st}^\star(X_s,X_t) 
  =&~ \alpha \Psi_{st}(X_s,X_t) \Psi_s(X_s) \Psi_t(X_t) \prod_{u\in \Gamma_s \setminus t} m_{us}^\star(X_s) \prod_{v\in \Gamma_t \setminus s} m_{vt}^\star(X_t), \\
  b_s^\star(X_s)
  =&~ \alpha \Psi_{s}(X_s) \prod_{u\in \Gamma_s } m_{us}^\star(X_s),
\end{align}
where $\alpha$ denotes a normalization constant and $\cbr{m_{ts}^\star}$ are the fixed point messages,
\begin{align}
  \label{eq:fixpoint_message}
  m_{ts}^\star(X_s) = \alpha \int_{\Xcal} \Psi_{st}(X_s,X_t) \Psi_t(X_t) \prod_{u\in \Gamma_t \setminus s} m_{ut}^\star(X_t)~dX_t. 
\end{align}
 Thus,
\begin{align}
  \PP^\star(X_s,X_t) 
  =&~ \alpha \Psi_{st}(X_s,X_t) \Psi_s(X_s) \Psi_t(X_t) \prod_{u\in \Gamma_s \setminus t} m_{us}^\star(X_s) \prod_{v\in \Gamma_t \setminus s} m_{vt}^\star(X_t), \label{eq:relation_marginal0}\\
  \PP^\star(X_s)
  =&~ \alpha \Psi_{s}(X_s) \prod_{u\in \Gamma_s } m_{us}^\star(X_s). \label{eq:relation_marginal}
\end{align}
Combining these relations and assuming that $\PP^\star(X_s)$ and $m_{ts}^\star(X_s)$ are strictly positive, we can also obtain the consistent relation for the local conditionals,
\begin{align}
  \label{eq:relation_conditional}
  \PP^\star(X_t | X_s) = \frac{\PP^\star(X_s,X_t)}{\PP^\star(X_s)} = \frac{\Psi_{st}(X_s,X_t) \Psi_s(X_t) \prod_{u\in \Gamma_t \setminus s} m_{ut}^\star(X_t)}{m_{ts}^\star(X_s)} .
\end{align}

\section{BP Inference Using Learned Potentials}\label{sec:BPusingLearnedPotentials_supp}

The inference problem in pairwise MRFs is to compute the marginals or the log partition function for the model with learned potentials. 
Belief propagation is an iterative algorithm for performing approximate inference in MRFs. BP can also be viewed as an iterative algorithm for minimizing the Bethe free energy approximation to the log partition function.  The results of this algorithm are a set of beliefs which can be used for obtaining the MAP assignment of the corresponding variables. 

The BP  message update (with the learned potentials) is
\begin{align}
  \label{eq:message_update}
  m_{ts}(X_s) = \alpha \int_{\Xcal} \Psi_{st}(X_s,X_t) \Psi_t(X_t) \prod_{u\in \Gamma_t \setminus s} m_{ut}(X_t)~dX_t, 
\end{align}
and at any iteration, the beliefs can be computed using the current messages,
\begin{align}
  \BB_{st}(X_s,X_t) 
  =&~ \alpha \Psi_{st}(X_s,X_t) \Psi_s(X_s) \Psi_t(X_t) \prod_{u\in \Gamma_s \setminus t} m_{us}(X_s) \prod_{v\in \Gamma_t \setminus s} m_{vt}(X_t), \\
  \BB_s(X_s)
  =&~ \alpha \Psi_{s}(X_s) \prod_{u\in \Gamma_s } m_{us}(X_s)  .
\end{align}
To see how the message update equation can be expressed using the true local conditional $\PP^\star(X_t | X_s)$, we divide both size of~\eq{eq:message_update}~by the fixed point message $m_{ts}^\star(X_s)$ during BP learning stage, and introduce $1 = \frac{\prod_{u\in\Gamma_t \setminus s} m^\star_{ut}(X_t)}{\prod_{u\in\Gamma_t \setminus s} m^\star_{ut}(X_t)}$. 
The message update equation in~\eq{eq:message_update} can then be re-written as
\begin{align}
  \frac{m_{ts}(X_s)}{m^\star_{ts}(X_s)} = 
  \int_{\Xcal} \frac{\Psi_{st}(X_s, X_t) \Psi_t(X_t) \prod_{u\in\Gamma_t \setminus s} m_{ut}(X_t)}{m^\star_{ts}(X_s)} \prod_{u \in \Gamma_t \setminus s} \frac{m_{ut}(X_t)}{m^\star_{ut}(X_t)} ~d X_t  .
\end{align}
The belief at any iteration becomes
\begin{align}
  \BB_s(X_s) = \alpha \Psi_s(X_s) \prod_{u\in\Gamma_s} m_{us}(X_s) = \rbr{\prod_{u\in\Gamma_s} \frac{m_{st}(X_s)}{m^\star_{st}(X_s)}}\rbr{\alpha \Psi_s(X_s) \prod_{u\in\Gamma_s} m^\star_{st}(X_s)}.
\end{align}
We reparameterize the message as
\begin{align}
  \label{eq:definition_fst}
  m_{st}(X_t) \longleftarrow \frac{m_{st}(X_t)}{m^\star_{st}(X_t)}~~.  
\end{align}
Since the potentials are learned via BP, we can use the relation in~\eq{eq:relation_conditional} to obtain 
\begin{align}
  m_{ts}(X_s) = \int_{\Xcal} \PP^\star(X_t|X_s) \prod_{u \in \Gamma_t \setminus s} m_{ut}(X_t) ~d X_t.
\end{align}
Similarly, we obtain from~\eq{eq:relation_marginal} that
\begin{align}
  \BB_s(X_s) = \rbr{\prod_{u\in\Gamma_s} m_{st}(X_s)} \PP^\star(X_s).
\end{align}

\paragraph{Messages from Evidence Node}
Given  evidence $x_t$ at node $X_t$, the outgoing message from $X_t$ to $X_s$ is
$m_{ts}(X_s) = \alpha \Psi_{st}(X_s, x_t)\Psi_t(x_t)$. Using similar reasoning to the case of an internal node, we have
\begin{align}  
  \frac{m_{ts}(X_s)}{m^\star_{ts}(X_s)} 
  &= \frac{\Psi_{st}(X_s, x_t)\Psi_t(x_t)}{m^\star_{ts}(X_s)}  \\
  &= \frac{\Psi_{st}(X_s, x_t)\Psi_t(x_t)\prod_{u\in\Gamma_t\setminus s} m^\star_{ut}(x_t)}{m^\star_{ts}(X_s)}\frac{1}{\prod_{u\in\Gamma_t\setminus s} m^\star_{ut}(x_t)} \\
  &= \PP^\star(x_t|X_s) \frac{1}{\prod_{u\in\Gamma_t\setminus s} m^\star_{ut}(x_t)} \\
  &\propto \PP^\star(x_t|X_s)
\end{align}
where  $\prod_{u\in\Gamma_t\setminus s} m^\star_{ut}(x_t)$ is  constant given a fixed value $X_t=x_t$.  Reparametrizing the message $m_{ts}(X_s)\leftarrow \frac{m_{ts}(X_s)}{m^\star_{ts}(X_s)}$,  the outgoing message from the evidence node is simply the true likelihood function evaluated at $x_t$.

\section{A Note on Kernelization of Gaussian BP}\label{sec:kernelGaussianBP_supp}

In this section, we consider the problem of defining a joint Gaussian graphical model in the feature space induced by a kernel. We follow
\cite{Bickson08} in our presentation of the original Gaussian BP setting. We will show that
assuming a Gaussian in an infinite feature space leads to challenges in interpretation and estimation of the model. 

Consider a pairwise MRF, 
\begin{equation}
\PP(\bm{X})=\prod_{s \in \Vcal}\Psi_{s}(X_s)\prod_{(s,t)\in\Ecal}\Psi_{st}(X_s,X_t).\label{eq:undirectedModel}\end{equation}
In the case of the Gaussian, the probability density function takes
the form
\begin{eqnarray*}
\PP(\bm{X}) & \propto & \exp\left(-\tfrac{1}{2}(\bf{X}-\mu)^{\top}A(X-\mu)\right)\\
 & \propto & \exp\left(-\tfrac{1}{2}\bf{X}^{\top}A\bf{X}-b^{\top}\bf{X}\right),\end{eqnarray*}
where $\bf{A}=\bf{C}^{-1}$ is the precision matrix, and 
\[\bf{A}\mu=\bf{b}.\]
Putting this in the form (\ref{eq:undirectedModel}), the node and
edge potentials are written
\begin{equation}
\Psi_{s}(X_s)\triangleq\exp\left(-\tfrac{1}{2}X_s^{\top}A_{ss}X_s +b_{s}X_s\right)\label{eq:nodePotential}\end{equation}
and
\begin{equation}
\Psi_{st}(X_s,X_t)\triangleq\exp\left(-X_s^{\top}A_{st}X_t\right).\label{eq:edgePotential}\end{equation}
We now consider how these operations would appear in Hilbert space.
In this case, we would have\[
\Psi_s(X_s):=\exp\left(-\tfrac{1}{2}\left\langle \phi(X_s),A_{ss}\phi(X_s)\right\rangle_{\Fcal} +\left\langle b_s,\phi(X_s)\right\rangle_{\Fcal} \right)\,,\]
and\[
\Psi_{st}(X_s,X_t):=\exp\left(-\left\langle \phi(X_s),A_{st}\phi(X_t)\right\rangle_{\Fcal} \right).\]
We call the $A_{ss}$ and $A_{st}$ \emph{precision operators}, by
analogy with the finite dimensional case. At this point, we already
encounter a potential difficulty in kernelizing Gaussian BP: how do
we learn the operators $A_{ss}$, $b_{s}$ and $A_{st}$ from data?
We could in principle define a covariance operator $\bf{C}$ with $(s,t)$th
block the pairwise covariance operator $C_{st}$, but $A_{st}$ would
then be the $(s,t)$th block of $\bf{C}^{-1}$, which is difficult to
compute. As we shall see below, however, these operators appear
in the BP message updates.

Next, we describe a message passing procedure for the Gaussian potentials
in (\ref{eq:nodePotential}) and (\ref{eq:edgePotential}). The message
from $t$ to $s$ is written 
\[
m_{ts}(X_s)=\int \limits_{\mathcal{X}}\Psi_{st}(X_s,X_t)\underset{(a)}{\underbrace{\Psi_t(X_t)\prod_{u\in \Gamma_t \setminus s}m_{ut}(X_t)}}~dX_t.\]
We first consider term (a) in the above. We will assume, with justification
to follow, that $m_{ut}(X_t)$ takes the form
\[
m_{ut}(X_t)\propto\exp\left(-\tfrac{1}{2}X_t^{\top}P_{ut}X_t+\mu_{ut}^{\top}X_t\right),\]
where the terms $P_{ut}$ and $\mu_{ut}$ are defined by recursions
specified below (we retain linear algebraic notation for simplicity). It follows that $\Psi_t(X_t)\prod_{u \in \Gamma_t \setminus s}m_{ut}(X_t)$
is proportional to a Gaussian,
\[
\Psi_t(X_t)\prod_{u \in \Gamma_t \setminus s}m_{ut}(X_t)\propto\exp\left(-\tfrac{1}{2}X_t^{\top}P_{t\backslash s}X_t+\mu_{t\setminus s}^{\top}X_t\right),\]
where we define the intermediate operators\[
\mu_{t\setminus s}:=\mu_t+\sum_{u\in \Gamma_t \setminus s}\mu_{ut}\]
and\[
P_{t\setminus s}:=A_{ss}+\sum_{u\in \Gamma_t \setminus s}P_{ut}\,.\]
To compute the message $m_{ts}(X_s)$, we integrate\begin{eqnarray}
m_{ts}(X_s)&=&\int \limits_{\mathcal{X}}\Psi_{st}(X_s,X_t)\Psi_t(X_t)\prod_{u\in \Gamma_t \setminus s}m_{ut}(X_t)~dX_t\nonumber \\
 & = & \int \limits_{\mathcal{X}}\exp\left(-X_tA_{ts}X_s \right)\exp\left(-\tfrac{1}{2}X_t^{\top}P_{t\backslash s}X_t+\mu_{t\setminus s}^{\top}X_t\right)~dX_t
 \label{eq:gaussianbp_update}
 \end{eqnarray}
Completing the square, we get the parameters of the message $m_{ts}$
in the standard form,
\begin{align}
P_{ts} &=-A_{ts}^{\top}P_{t\setminus s}^{-1}A_{ts} \label{eq:gaussianparam_update1}\\ 
\mu_{ts} &= -\mu_{t\setminus s}^{\top}P_{t\setminus s}^{-1}A_{ts}.  \label{eq:gaussianparam_update2}
\end{align}

There are two main difficulties in implementing the above procedure in feature space. First,  it is not clear how to learn the precision operators from the data. Second, we need to invert these precision operators. Thus, it remains a challenging open
question to define Gaussian BP in feature space. The feature space Gaussian BP updates may be contrasted with the kernel BP updates we propose in the main text. The latter have regularized closed form empirical estimates, and they are very different from the Gaussian BP form in~\eq{eq:gaussianbp_update} and parameter updates in~\eq{eq:gaussianparam_update1}~and~\eq{eq:gaussianparam_update2}.

\section{Message Error Incurred by the Additional Feature Approximation}\label{sec:messageError_supp}

We  bound the difference between the estimated conditional embedding operator $\hUcal_{X_t^\otimes|X_s}$ and its counterpart $\tUcal_{X_t^\otimes|X_s}$ after further feature approximation. 
Assume $\nbr{\phi(x)}_{\Fcal} \leq 1$, and define the tensor feature $\xi(x) := \bigotimes_{u\setminus s} \phi(x)$. Denote by $\tilde{\xi}(x)$ and $\tilde{\phi}(x)$ the respective
approximations of $\xi(x)$ and $\phi$. 
Furthermore, let the approximation error after the incomplete QR decomposition be $\epsilon= \max\cbr{\max_{\Xcal} \nbr{\phi(x) - \tilde{\phi}(x)}_{\Fcal},~\max_{\Xcal} \nbr{\xi(X) - \tilde{\xi}(x)}_{\Hcal}}$. It follows that 
\begin{align}
  &\nbr{\hUcal_{X_t^\otimes|X_s} - \tUcal_{X_t^\otimes|X_s}}_{HS} \\
  \leq&~ \nbr{
  \hCcal_{X_t^\otimes X_s} (\hCcal_{X_s X_s} + \lambda_m I)^{-1} -
  \tCcal_{X_t^\otimes X_s} (\tCcal_{X_s X_s} + \lambda_m I)^{-1} 
  }_{HS} \\
  \leq&~ \nbr{(\hCcal_{X_t^\otimes X_s}-\tCcal_{X_t^\otimes X_s})(\hCcal_{X_s X_s} + \lambda_m I)^{-1}}_{HS} 
  + \nbr{\tCcal_{X_t^\otimes X_s}\sbr{(\tCcal_{X_sX_s} + \lambda_m I)^{-1} - (\hCcal_{X_s X_s} + \lambda_m I)^{-1}} }_{HS} \\
  \leq&~ \frac{1}{\lambda_m} \nbr{\hCcal_{X_t^\otimes X_s}-\tCcal_{X_t^\otimes X_s}}_{HS} + \frac{1}{\lambda_m^{3/2}} \nbr{\hCcal_{X_s X_s} - \tCcal_{X_s X_s}}_{HS}.
\end{align}
For the first term,
\begin{align} 
  &\frac{1}{\lambda_m} \nbr{\hCcal_{X_t^\otimes X_s}-\tCcal_{X_t^\otimes X_s}}_{HS} \\
  =&~ \frac{1}{\lambda_m} \nbr{\frac{1}{m}\sum_{i} \xi(x_s^{i}) \phi(x_s^{i})^\top - \frac{1}{m}\sum_{i} \tilde{\xi}(x_s^{i}) \tilde{\phi}(x_s^{i})^\top }_{HS} \\
  \leq&~ \frac{1}{\lambda_m} \frac{1}{m}\sum_{i} \nbr{\xi(x_s^{i}) \phi(x_s^{i})^\top - \tilde{\xi}(x_s^{i}) \tilde{\phi}(x_s^{i})^\top }_{HS} \\
  \leq&~\frac{1}{\lambda_m} \max_i \nbr{\xi(x_s^{i}) \phi(x_s^{i})^\top - \tilde{\xi}(x_s^{i}) \tilde{\phi}(x_s^{i})^\top }_{HS} \\
  \leq&~\frac{1}{\lambda_m} \max_i \cbr{\nbr{\xi(x_s^{i}) \phi(x_s^{i})^\top - \tilde{\xi}(x_s^{i}) \phi(x_s^{i})^\top }_{HS} + \nbr{\tilde{\xi}(x_s^{i}) \phi(x_s^{i})^\top - \tilde{\xi}(x_s^{i}) \tilde{\phi}(x_s^{i})^\top }_{HS}}\\
  \leq&~ \frac{2 \epsilon}{\lambda_m}  . 
\end{align}
Similarly, for the second term, 
\begin{align}
  \frac{1}{\lambda_m^{3/2}} \nbr{\hCcal_{X_s X_s} - \tCcal_{X_s X_s}}_{HS} \leq \frac{2 \epsilon}{\lambda_m^{3/2}}.
\end{align}
Combining the results, we obtain
\begin{align}
  \nbr{\hUcal_{X_t^\otimes|X_s} - \tUcal_{X_t^\otimes|X_s}}_{HS} \leq 2\epsilon(\lambda_m^{-1} + \lambda_m^{-3/2}).
\end{align}

\bibliographystyle{natmlapa}
 \bibliography{bibfile,danny-extra-bib}

\end{document}